\documentclass{article}
\PassOptionsToPackage{numbers, compress}{natbib}

\usepackage[preprint]{neurips_2021}

\usepackage[utf8]{inputenc} 
\usepackage[T1]{fontenc}    
\usepackage{hyperref}       
\usepackage{url}            
\usepackage{booktabs}       
\usepackage{amsfonts}       
\usepackage{nicefrac}       
\usepackage{microtype}      
\usepackage{xcolor}         

\usepackage{grffile}

\usepackage{amssymb}
\usepackage{bibentry}
\usepackage{amsfonts}
\usepackage{subfigure}
\usepackage{multirow}
\usepackage{amsmath}
\usepackage{graphicx}
\usepackage{epsfig}
\usepackage{color}
\usepackage{epstopdf}
\usepackage{rotating}
\usepackage{caption}
\usepackage{float}
\usepackage{multicol}
\usepackage{amssymb}
\usepackage{bbding}
\usepackage{balance}
\usepackage{color}

\usepackage{float}
\usepackage{stfloats}

\usepackage{adjustbox}
\usepackage{algorithm,algorithmicx}
\usepackage{algpseudocode}

\newcommand{\argmax}{\operatornamewithlimits{arg\,max}}

\usepackage{balance}

\usepackage{wrapfig}
\usepackage{lipsum}

\DeclareMathAlphabet\mathbfcal{OMS}{cmsy}{b}{n}

\usepackage{threeparttable,color}
\usepackage{booktabs}

\newcommand{\explainer}{\textsc{GnnExplainer}}
\newcommand{\ourname}{{GEAttack}}

\newcommand{\gray}[1]{{\color{gray}~#1 }}

\title{Jointly Attacking Graph Neural Network and its Explanations}

\author{%
Wenqi Fan$^1$\quad Wei Jin$^2$\quad Xiaorui Liu$^2$\quad Han Xu$^2$\quad Xianfeng Tang$^3$\quad \\\textbf{Suhang Wang}$^3$\quad  \textbf{Qing Li}$^1$\quad  \textbf{Jiliang Tang}$^2$\quad  \textbf{Jianping Wang}$^4$\quad  \textbf{Charu Aggarwal}$^5$\\
$^1$The Hong Kong Polytechnic University\quad $^2$Michigan State University\\
$^3$The Pennsylvania State University\quad $^4$City University of Hong Kong\quad $^5$IBM T.J. Watson\\
\texttt{wenqifan03@gmail.com}\quad 
\texttt{$^2$\{jinwei2, xiaorui, xuhan1, tangjili\}@msu.edu}\\
\texttt{$^3$\{xut10, szw494\}@psu.edu}\quad 
\texttt{csqli@comp.polyu.edu.hk}\quad \\
\texttt{jianwang@cityu.edu.hk}\quad 
\texttt{charu@us.ibm.com}
}
 
\begin{document}

\maketitle

\begin{abstract}
 Graph Neural Networks (GNNs) have boosted the performance for many graph-related tasks. Despite the great success, recent studies have shown that GNNs are highly vulnerable to adversarial attacks, where adversaries can mislead the GNNs' prediction by modifying graphs. On the other hand, the explanation of GNNs ({\explainer}) provides a better understanding of a trained GNN model by generating a small subgraph and features that are most influential for its prediction. In this paper, we first perform empirical studies to validate that {\explainer}  can act as an inspection tool and have the potential to detect the adversarial perturbations for graphs. This finding motivates us to further initiate a new problem investigation: \textit{Whether a graph neural network and its explanations can be jointly attacked by modifying graphs with malicious desires?} 
It is challenging to answer this question since the goals of adversarial attacks and bypassing the {\explainer} essentially contradict each other.
In this work, we give a confirmative answer to this question by proposing a novel attack framework (\textbf{GEAttack}), which can attack both a GNN model and its explanations by simultaneously exploiting their vulnerabilities. 
Extensive experiments on two explainers ({\explainer} and PGExplainer) under various real-world datasets demonstrate the effectiveness of the proposed method.
 
\end{abstract}


\section{Introduction}

Graph neural networks (GNNs) have achieved significant success for graphs in various real-world applications~\cite{kipf2016semi,derr2020epidemic,fan2020graph},  such as node classification~\cite{hamilton2017inductive,jin2021node}, recommender systems~\cite{fan2019graph,fan2019deep_daso,fan2019deep_dscf,fan2018deep}, and natural language processing~\cite{liu2019does,xu2019adversarial,beck2018graph}. 
Despite the great success, recent studies show that GNNs are highly vulnerable to adversarial attacks~\cite{zugner2018adversarial,dai2018adversarial,wang2019attacking}, which has raised great concerns for employing GNNs in security-critical applications~\cite{xu2019adversarial,dai2018adversarial,sun2020adversarial}. 
More specifically, attackers can insert adversarial perturbations into graphs, which can lead a well-designed  model to produce incorrect outputs or have bad overall performance~\cite{wu2019adversarial,xu2019adversarial,xu2019topology}. 
For example, adversaries can build well-designed user profiles to promote/demote items in bipartite graphs on many e-commerce platforms such as Alibaba and  Amazon~\cite{fan2020attacking}; or hackers might intend to damage the reputations of an elector's main opponents by propagating fake news in social media~\cite{sun2020adversarial}.

\begin{figure*}[htb]
\centering
\includegraphics[height=70mm, width=130mm]{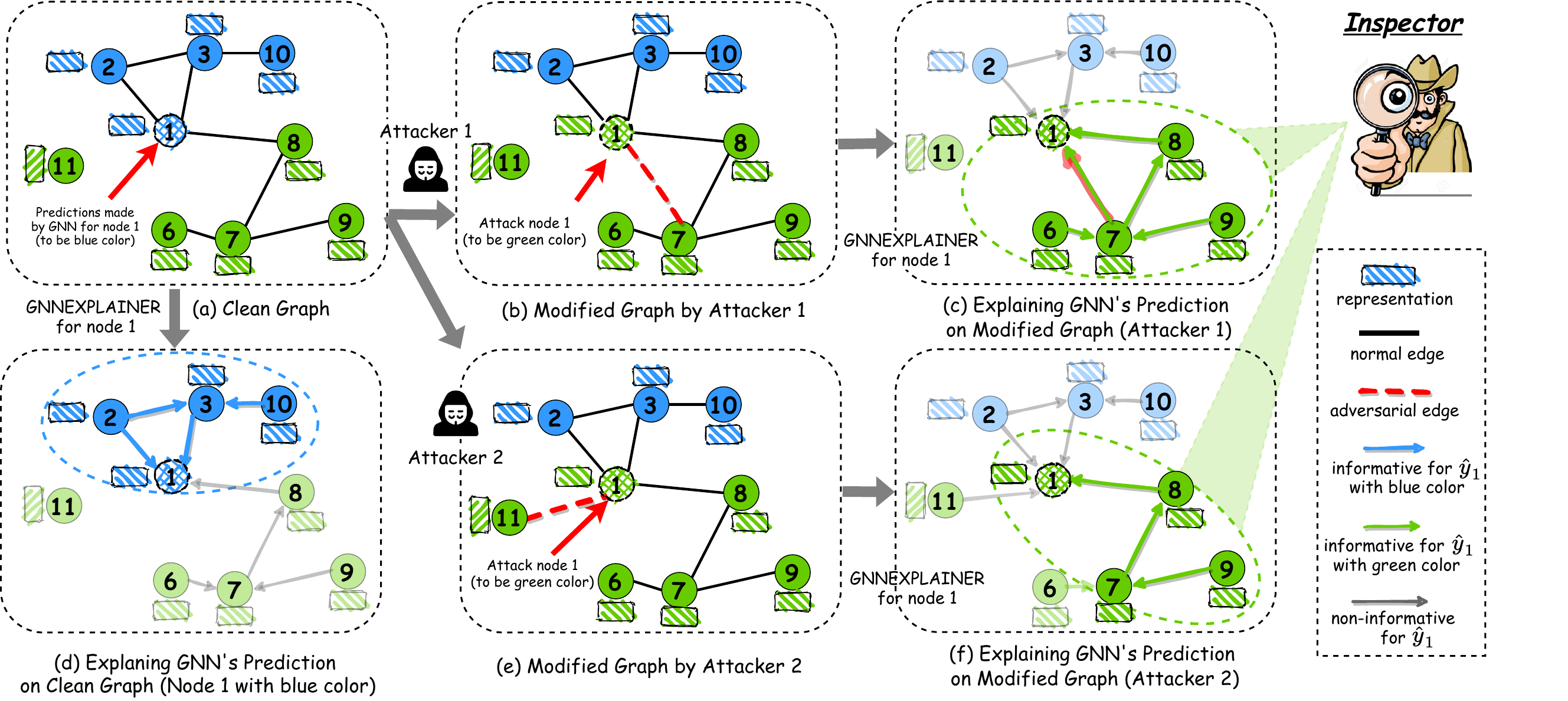}
\caption{Adversarial attacks and the explanations ({\explainer}) for prediction made by a GNN model. Some edges form important message-passing pathways (in dotted circle with blue/green color) while others do not (in translucent). The \emph{Attacker $1$} can successfully change the GNN's prediction to green color on the target node $v_1$, while the added adversarial edge $(v_1, v_7)$ is included into a subgraph generated by {\explainer}. While the \emph{Attacker $2$} can  attack the GNN model, as well as fool the {\explainer}, where  the added adversarial edge $(v_1, v_{11})$ is not included into a subgraph and successfully evade the detection by an inspector.}
\label{fig:intro_explainer}
\end{figure*}

Recently, interpretation methods for GNNs~\cite{ying2019gnnexplainer,yuan2020xgnn,huang2020graphlime,luo2020parameterized} have been proposed to explain the inner working mechanisms of GNNs. In particular, given a trained GNN model and its prediction on a test node,  {\explainer}~\cite{ying2019gnnexplainer, luo2020parameterized} will return a small subgraph together with a small subset of node features that are most influential for its prediction. On account of this, the model's decision via the output subgraphs and features can be well-interpreted by people.
In our work, we hypothesize that {\explainer} can also provide great opportunities for people (e.g., inspectors or system designers)  to inspect the "confused" predictions from adversarial perturbations. For example, in the e-commerce platform systems, once there exist abnormal predictions for certain products given by the GNN model,  \explainer~can help locate the most influential users leading the model to make such a prediction. In this way, we can figure out the abnormal users, or adversaries who deliberately propagate such adversarial information to our system. 
As an illustrative example in Figure~\ref{fig:intro_explainer}, an attacker (\emph{Attacker $1$}) changes node $v_1$'s prediction of the GNN model by adding an adversarial edge $(v_1, v_7)$ (cf. Figure~\ref{fig:intro_explainer} (b)). At this time, if people find that $v_i$'s prediction is problematic, they can apply graph interpretation methods (cf. Figure~\ref{fig:intro_explainer} (c)) to ``inspect'' this anomaly. Note that {\explainer} can figure out the most influential components (a
In such case, the adversarial edges made by attackers are highly likely to be chosen by {\explainer} and then detected by the inspector or system designers (cf. Figure~\ref{fig:intro_explainer} (c)). To this end, adversarial users/edges can be excluded from the GNN model to improve the system's safety. In fact, this hypothesis is verified via empirical studies on real-world datasets. The details can be found in Section~\ref{sec:pre}.

Motivated by the fact that {\explainer} can act as an inspection tool for graph adversarial perturbations, we further investigate a new problem: \textit{Whether a graph neural network and its explanations can be jointly attacked by modifying graphs with malicious desires?} For example, as shown in Figure~\ref{fig:intro_explainer} (e-f), when an attacker (Attacker 2) inserts an adversarial edge $(v_1, v_{11})$ to mislead the model's prediction, he can also successfully evade the detection by a \explainer. In such a scenario, the attacker becomes more dangerous where he can even misguide these inspection approaches, leading to more severe safety issues for GNNs. However, jointly fooling graph neural networks and its explanations faces tremendous challenges. The biggest obstacle is that the goals of adversarial attacks and bypassing the {\explainer} essentially contradict each other. After all, the adversarial perturbations on graphs are highly correlated with the target label, since it is the perturbations that cause such malicious prediction~\cite{ying2019gnnexplainer,luo2020parameterized}. Therefore, the {\explainer} has a high chance to detect these perturbations by maximizing the mutual information with GNN's prediction. Moreover, although there exist extensive works on adversarial attack for a GNN model~\cite{zugner2018adversarial,dai2018adversarial,jin2020adversarial,zugner_adversarial_2019,sun2020adversarial,wu2019adversarial}, the joint attack on the GNN model and its explanations is rarely studied which calls for new solutions.  
 
To address the aforementioned challenges, in this work, we propose a novel attack framework~(\textbf{\ourname}) for graphs, where an attacker can successfully fool the GNN model and misguide the inspection from {\explainer} simultaneously. In particular, we first validate that \explainer ~tools can be utilized to understand and inspect the problematic outputs from adversarially perturbed graph data, which paves a way to improve the safety of GNNs. After that, we propose a new attacking problem, where we seek to jointly attack a graph neural network method and its explanations. Our proposed algorithm \textbf{GEAttack} successfully resolves the dilemma between attacking a GNN and its explanations by exploiting their vulnerabilities simultaneously. To the best of our knowledge, we are the first to study this problem that reveals more severe safety concerns.
Experimental results on two explainers ({\explainer}\cite{ying2019gnnexplainer} and PGExplainer~\cite{luo2020parameterized}) demonstrate that {\ourname} achieves good performance for attacking GNN models, and adversarial edges generated by \ourname~ are much harder to be detected by {\explainer}, which successfully achieve the joint attacks on a GNN model and its explanations.

 \section{Related Work}\label{related}

\textbf{Adversarial Attacks on Graphs.} GNNs generalize deep neural networks to graph structured data and become powerful tools for graph representations learning~\cite{jin2020self,kipf2016semi,liu2021elastic}. However, recent studies have demonstrated that GNNs suffer from the same issue as other deep neural networks: they are highly vulnerable to adversarial attacks~\cite{liu2021trustworthy,dai2018adversarial,xu2019adversarial,zugner_adversarial_2019,ma2020towards,chang2020restricted}. 
Specifically, attackers can generate graph adversarial perturbations by manipulating the graph structure or node features to deceive the GNN model for making incorrect predictions~\cite{zugner2018adversarial,dai2018adversarial,xu2019adversarial}. 
\textit{Nettack}~\cite{zugner2018adversarial} is one of the first methods that perturbs the graph structure data by preserving  degree distribution and feature co-occurrence to perform attack on GNN model~\cite{kipf2016semi}.  \textit{RL-S2V}~\cite{dai2018adversarial} is the first work to employ reinforcement learning to generate adversarial perturbations on graph data. 
\textit{NIPA}~\cite{sun2020adversarial} also proposes a deep reinforcement learning based method to perform fake node injection attack on graph by simulating the attack process, and sequentially adds the adversarial edges and designs labels for the injected fake nodes.  
\textit{IG-Attack}~\cite{wu2019adversarial} introduces an integrated gradients based attack method to accurately reflect the effect of perturbing certain features or edges on graph data. \textit{Metattack}~\cite{zugner_adversarial_2019} is proposed to globally perturb the graph based on meta learning technique. 
In our work, we first claim that {\explainer} tools can serve as an alternative way to improve the GNN model's safety, by people (such as system inspectors or designers) doing inspections on the problematic prediction outcomes of GNN models, and then locating the potential adversarial perturbations in the graph.

\textbf{Explaining GNNs.} The explanation techniques of deep models aim to study the underlying relationships behind the predictions of deep models, and provide human-intelligible explanations, which can make the deep models more trustable~\cite{ribeiro2016should,vu2020pgm,yuan2020xgnn}.  Some recent efforts have been made to explain the deep models for image and text data~\cite{selvaraju2017grad,zhou2016learning,ribeiro2016should,kim2020interpretation}. However, the explainability of graph neural network models on graph structured data is less explored, which is critical for understanding deep graph neural networks ~\cite{ying2019gnnexplainer,yuan2020xgnn,huang2020graphlime,luo2020parameterized,yuan2021explainability}.  In particular, as one of the very first methods to interpret GNN, {\explainer}~\cite{ying2019gnnexplainer} maximizes the mutual information between the distribution of possible subgraphs and the GNN’s prediction to find the subgraph that is most influential for the prediction. PGExplainer~\cite{luo2020parameterized} is proposed to generate an explanation for each instance with a global understanding of the target GNN model in an inductive setting.
GraphLIME~\cite{huang2020graphlime} proposes a local interpretable explanation framework for GNN with the Hilbert-Schmidt Independence Criterion (HSIC) Lasso.
Meanwhile, in order to investigate what input patterns can result in a certain prediction, XGNN~\cite{yuan2020xgnn} is proposed to train a graph generator for finding graph patterns to maximize a certain prediction of the target GNN model by formulating the graph generation process as a reinforcement learning problem. The improved interpretability is believed to offer a sense of security by involving human in the decision-making process~\cite{yuan2020xgnn,ribeiro2016should}. However,  given its data-driven nature,  the interpretability itself is potentially susceptible to malicious manipulations~\cite{xu2019adversarial}. 
Note that there are other efforts devoted to connecting these two topics by attacking interpretation methods~\cite{liu2018adversarial,ghorbani2019interpretation,heo2019fooling,zhang2020interpretable} on non graph-structured data. 
In this work, our goal is to fool a GNN model as well as its interpretation methods. 
To the best of our knowledge,  this is the very first effort to attack both GNNs model and its explanations on graph structured data.

\section{Preliminary Study}\label{sec:pre}
 
In this section, we investigate the potential of {\explainer}~\cite{ying2019gnnexplainer} to detect adversarial attacks through empirical study. Note that in the remaining of the paper, we focus on {\explainer}; {however, the proposed methods can be also applied to other explanation methods and detailed results can be found in the Section~\ref{sec:jointly_pge}.} Next, we first introduce key notations and concepts used in this work.

\textbf{Graph Neural Networks.} Formally, let $G = (V, E)$ denote a graph where $V =\{v_1, ..., v_n\}$ is the set of $n$ nodes and $E = \{e_1, ..., e_k\}$ is the edge set. We use $\mathbf{A} \in \{0,1\}^{n\times n}$ to indicate the adjacency matrix of $G$, where the $i,j$-th element ${A}_{ij}$ is 1 if node $v_i$ and node $v_j$ are connected in $G$, and 0 otherwise. We also use  $\mathbf{X} =[\mathbf{x}_1,\mathbf{x}_2, ..., \mathbf{x}_n] \in \mathbb{R}^{n\times d}$ to represent the node feature matrix, where $\mathbf{x}_i$ is the $d$-dimensional feature vector of the node $v_i$. Here we use $G = (\mathbf{A}, \mathbf{X})$ to represent the graph data. Without loss of generality, given a graph $G = (\mathbf{A}, \mathbf{X})$,  we consider the problem of node classification for GNNs to learn a function $f_\theta: V_L \rightarrow Y_L$, that maps a part of nodes $V_L=\left \{ v_1, v_2, ..., v_l \right \}$ to their corresponding labels $Y_L=\left \{ y_1, y_2, ..., y_l \right \}$~\cite{kipf2016semi}. 
The objective function of GNNs can be defined as:
{\small
\begin{align}
\min _{\theta}~ \mathcal{L}_{\text{GNN}}(f_\theta(\mathbf{A}, \mathbf{X})) &:=\sum_{v_i \in {V}_{L}} \ell \left(f_{\theta}(\mathbf{A}, \mathbf{X})_{v_i}, y_{i}\right)  
=- \sum_{v_i \in V_L} \sum_{c=1}^C \mathbb{I}[ y_i= c ]\ln (f_{\theta}(\hat{\mathbf{A}}, \mathbf{X})_{v_{i}}^{c} ).
\end{align}
}  
where $\theta$ is the parameters of a GNN model $f_{\theta}$. 
$f_\theta( { \mathbf{A}, \mathbf{X}})_{v_i}^{c}$ denotes the $c$-th softmax output of node $v_i$ and $C$ is the number of class. $y_i$ is the true label of node $v_i$ and $\ell(\cdot, \cdot)$ is the cross-entropy loss function.
In this work, we adopt a two-layer GCN model~\cite{kipf2016semi} with $\theta=(\mathbf{W}_1, \mathbf{W}_2)$ as:
$ f_\theta(\mathbf{A}, \mathbf{X}) = \operatorname{softmax}(\tilde{\mathbf{A}} \thinspace \sigma( \tilde{\mathbf{A}} \thinspace \mathbf{X}\thinspace \mathbf{W}_1)  \thinspace \mathbf{W}_2)$,
where $\tilde{\bf A}=\tilde{\bf D}^{-1 / 2}({\bf A}+{\bf I}) \tilde{\bf D}^{-1 / 2}$ and $\tilde{\bf D}$ is the diagonal matrix of ${\bf A} + {\bf I}$ with $\tilde{\bf D}_{ii} = 1 + \sum_{j} {\bf A}_{ij}$. $\operatorname{\sigma}$ is the activation function such as ReLU.

\begin{wrapfigure}{R}{0.50\textwidth}
\centering
    {\subfigure[CITESEER]
    {\includegraphics[width=0.492\linewidth]{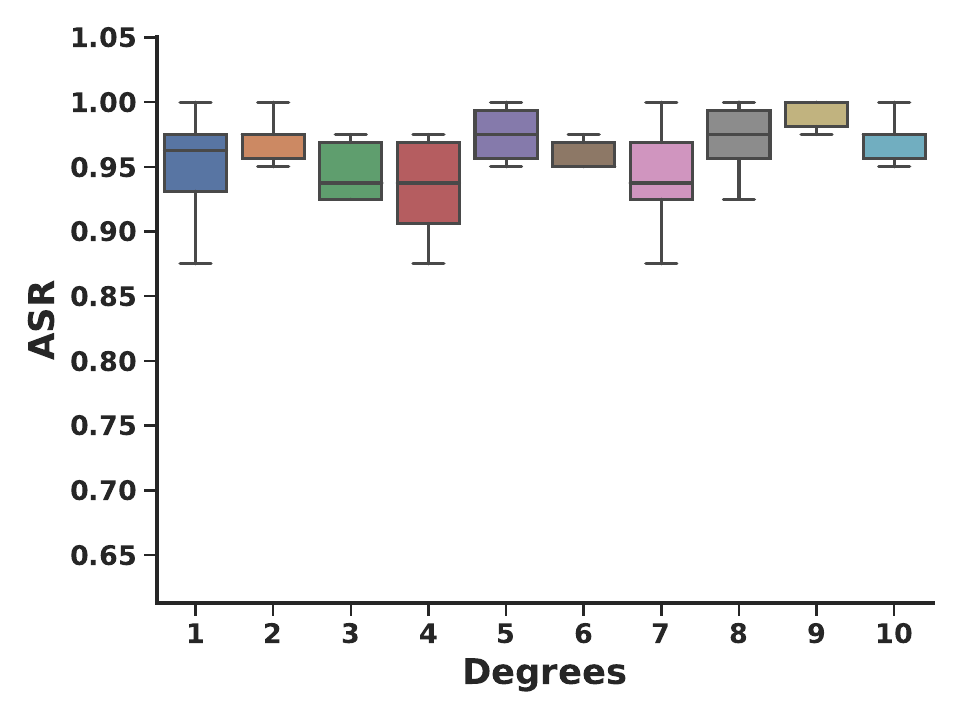}}}
    {\subfigure[CORA]
    {\includegraphics[width=0.492\linewidth]{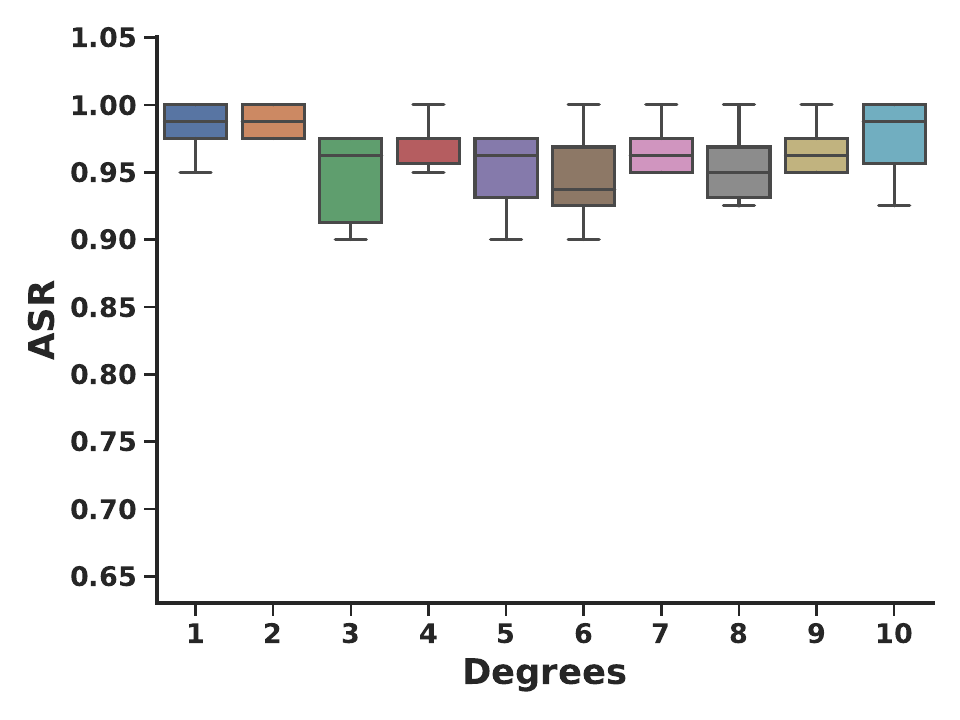}}}
    \caption{Results of Attack Success Rate (ASR) under Nettack on CITESEER and CORA datasets.} \label{fig:preliminarries_main}
\end{wrapfigure}

\textbf{{\explainer}.} In order to explain why a GNN model $f_\theta$ predicts a given node $v_i$'s label as $Y$, the {\explainer}  acts to provide a local interpretation  $G_S= (\mathbf{A}_S, \mathbf{X}_S)$  by highlighting the relevant features $\mathbf{X}_S$ and the relevant subgraph structure $\mathbf{A}_S$ for its prediction~\cite{ying2019gnnexplainer,luo2020parameterized}. To achieve this goal, it formalizes the problem as an optimization task to find the optimal explanation ($G_S$), which has the maximum {Mutual Information} (MI) with the GNN's prediction $Y$: $ MI\left(Y, (\mathbf{A}_S, \mathbf{X}_S)\right) := H(Y) - H(Y | \mathbf{A}= \mathbf{A}_S, \mathbf{X}=\mathbf{X}_S)$.
As the GNN model $f_\theta$ is fixed, the entropy term $H(Y)$ is also fixed.
In other words, the explanation for $v_i$'s prediction $\hat{y}_i$ is a subgraph $\mathbf{A}_S$ and  associated features $\mathbf{X}_S$  that minimize  the uncertainty of the GNN $f_\theta$ when the neural message-passing is limited to $G_S$ as:  
$\max_{(\mathbf{A}_S, \mathbf{X}_S)} MI\left(Y, (\mathbf{A}_S, \mathbf{X}_S)\right)
    \approx \min_{(\mathbf{A}_S, \mathbf{X}_S)}  - \sum_{c =1}^C \mathbb{I}[ \hat{y}_i=c ] \ln f_{\theta} (\mathbf{A}_S, \mathbf{X}_S)_{v_i}^c.$

Experimentally, the objective function of {\explainer} can be optimized to learn adjacency mask matrix $\mathbf{M}_{A}$ and feature selection mask matrix $\mathbf{M}_{F}$ in the following manner~\cite{ying2019gnnexplainer,luo2020parameterized}:
{\small
\begin{align}
    \min_{(\mathbf{M}_{A}, \mathbf{M}_{F})}  \mathcal{L}_{\text{Explainer}}(f_{\theta}, \mathbf{A}, \mathbf{M}_A, \mathbf{X}, \mathbf{M}_F, v_i, \hat y_i) 
    := -\sum_{c =1}^C \mathbb{I}[ \hat{y}_i=c ] \ln f_{\theta} 
    (\mathbf{A} \odot \sigma(\mathbf{M}_{A}), \mathbf{X} \odot  \sigma(\mathbf{M}_{F}))_{v_i}^c  
\label{eq:explainer_loss}
\end{align}
} 
where $\odot$ denotes element-wise multiplication, and $\sigma$ is the sigmoid function that maps the mask to $[0, 1]^{^{n\times n}}$.
After the optimal mask $\mathbf{M}_{A}$ is obtained, we can compute $\mathbf{A}_S = \mathbf{A} \odot \sigma (\mathbf{M}_{A})$  and use a threshold to remove low values. 
Finally, top-$L$ edges with the largest values can provide an explanation $\mathbf{A}_S$ for GNN’s prediction at node $v_i$. The same operation $\mathbf{X}_S =\mathbf{X} \odot \sigma(\mathbf{M}_{F})$   can be used to produce explanations by considering the feature information.

\begin{figure}[t]
\centering
{\subfigure[CITESEER-F1]
{\includegraphics[width=0.240822\linewidth]{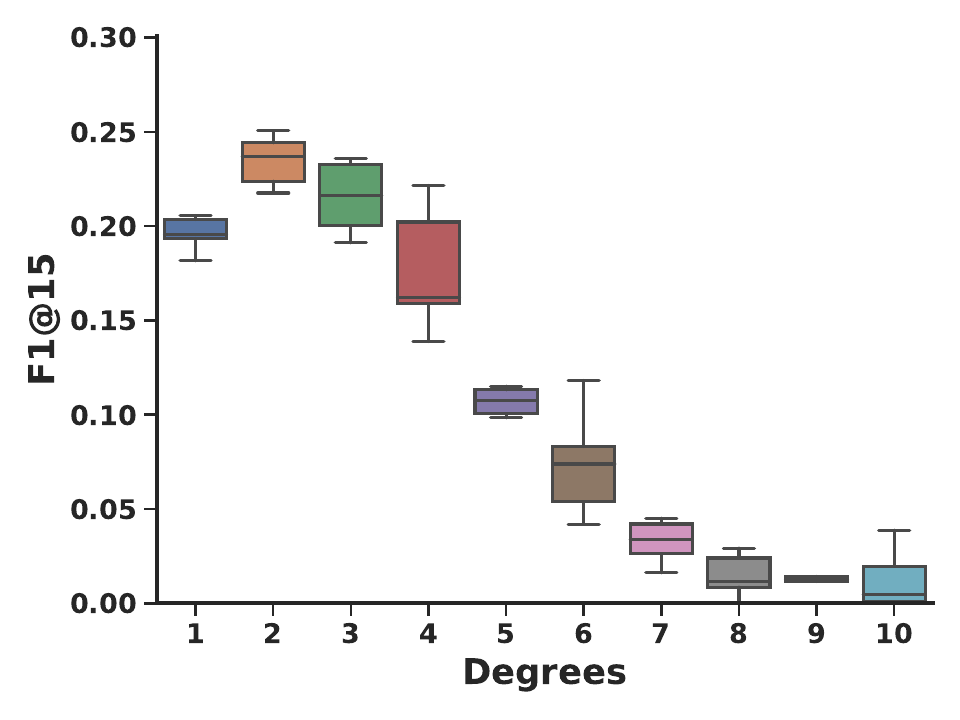}}}
{\subfigure[CITESEER-NDCG]
{\includegraphics[width=0.240822\linewidth]{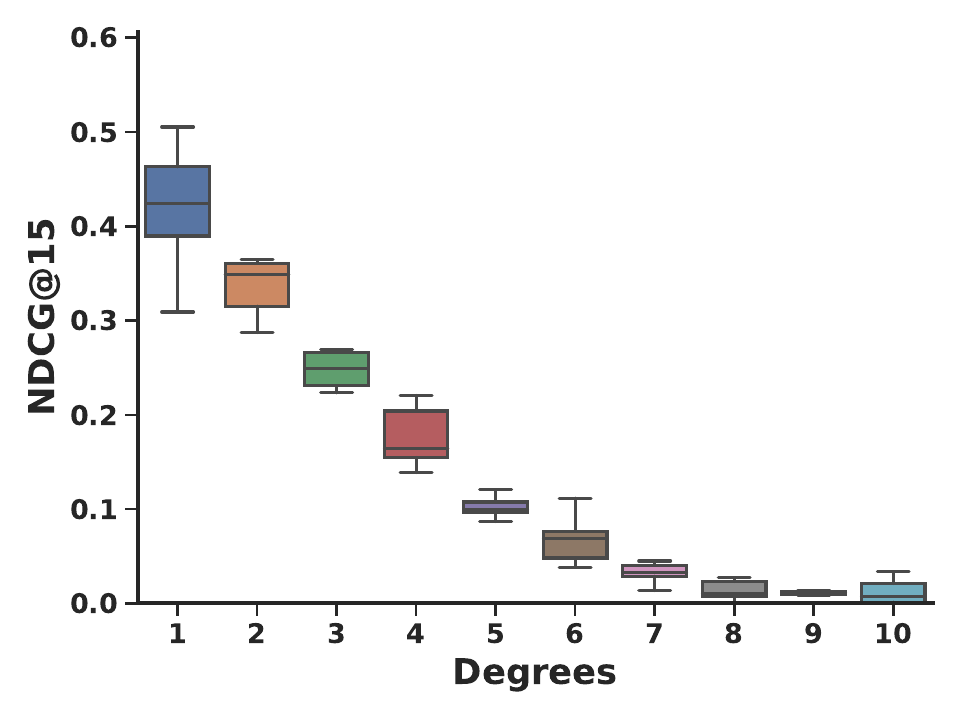}}}
{\subfigure[CORA-F1]
{\includegraphics[width=0.240822\linewidth]{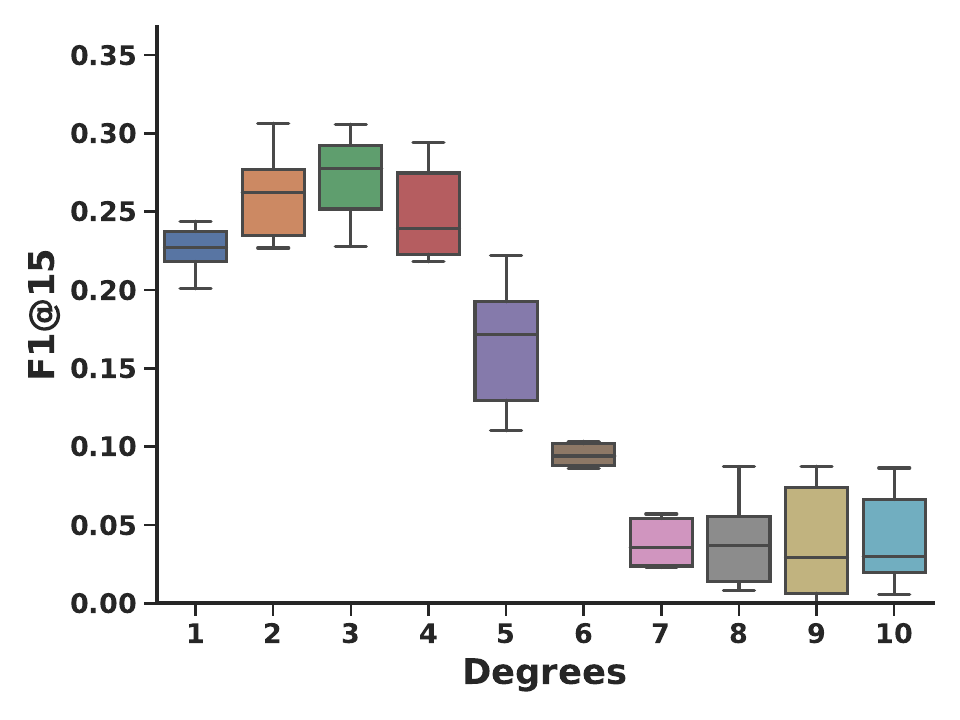}}}
{\subfigure[CORA-NDCG]
{\includegraphics[width=0.240822\linewidth]{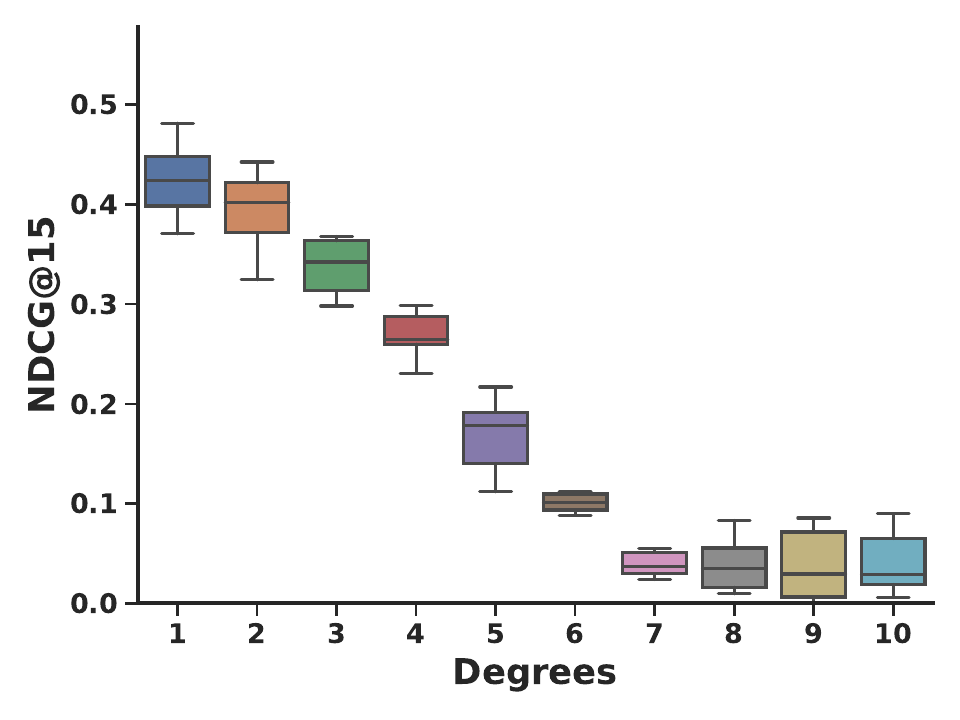}}}
\caption{Results of detecting the adversarial edges via {{\explainer}}   under Nettack on CITESEER and CORA datasets.}\label{fig:preliminarries_main_degrees}
\end{figure}


\textbf{{\explainer} as Adversarial Inspector.} In our work, we first hypothesize that if a model gives a wrong prediction to a test node $v_i$ because of some adversarially inserted fake edges, these ``adversarial edges'' should make a great contribution to the model's prediction outcome. Therefore, if a \explainer ~ can understand this wrong prediction outcome by figuring out the most influential edges, we are highly likely to find and locate the adversarial edges and finally exclude them from data. 
In particular, {\explainer} can reduce search space by generating a subgraph, and then domain experts can efficiently inspect the anomaly in the subgraph. 
For example, in a GNN based system for credit card fraud detection, when the system predicts a high-risk transaction for a consumer, financial experts can adopt a  {\explainer} to generate a small subgraph with some influential nodes (factors/features) from millions of features for this transaction. Then, financial experts use their professional knowledge to effectively inspect the anomaly in the generated subgraph.

In this subsection, we first dispatch empirical studies to validate the above mentioned hypothesis.
Note that in this work we concentrate on GNN's explanation on graph structural adversarial perturbations (for graph feature perturbations are similar, we leave it for future work).
Thus, the objective function of {\explainer} for adversarial edges detection can be defined as:
{\small
\begin{align}
    \min_{\mathbf{M}_{A}}  \mathcal{L}_{\text{Explainer}}(f_{\theta}, \mathbf{A}, \mathbf{M}_A, \mathbf{X}, v_i, \hat y_i)   
    \rightarrow \max_{\mathbf{M}_{A}}   \sum_{c =1}^C \mathbb{I}[ \hat {y}_i = c] \ln f_{\theta} (\mathbf{A} \odot \sigma(\mathbf{M}_{A}), \mathbf{X})_{v_i}^{c}.
    \label{equ: l_explainer_edges}
\end{align}
}\vspace*{-0.15\baselineskip}
where we find an optimal adjacency mask matrix $\textbf{M}_{A}$, namely the influential subgraph for $v_i$'s prediction. 
To check the inspection performance and verify our hypothesis, we check whether 
the influential subgraph generated by {\explainer} can help detect the adversarially inserted fake edges.

\noindent\textbf{Inspection Performance.} We conduct preliminary experiments on {two real-world datasets (i.e., CITESEER and CORA). }
The details of experimental settings can be found in Section~\ref{sec:experiments_setting}.
In these experiments, we choose the state-of-the-art graph attack method for GNN model, \textit{Nettack}~\cite{zugner2018adversarial}, to perturb the graph data by adding adversarial edges only\footnote{Removing edges or modifying nodes' features is much expensive and harder than adding fake connections among nodes in real-world social networks such as Facebook and Twitter~\cite{xu2020attacking,sun2020adversarial}\label{footnode:adding_edges}.}.
For each different node’s degree, we randomly choose 40 target (victim) nodes to validate whether the adversarial edges can be found by {\explainer} for them.  To extract the subgraph ($G_S$) for explanation, {\explainer} first computes the important weights  on edges  via masked adjacency ($\mathbf{M}_{A}$), and then uses a threshold to remove the edges with low values. Finally, top $L$ edges provide the explanation ($G_S$) for the GNN's prediction at target node~\cite{ying2019gnnexplainer}. In other words, adversarial edges with higher important weights are more likely to present at top ranks and be easily detected by people (such as system inspectors or designers).
Here, we adopt the F1 and NDCG to evaluate the detection rate on adversarial edges, where higher values of F1 and NDCG indicate that the adversarial edges are more likely to be detected and noticeable. 
As shown in Figures~\ref{fig:preliminarries_main} and~\ref{fig:preliminarries_main_degrees}, the \textit{Nettack} attacker can perform effective attacks for nodes with different degrees, achieving around 95\% attack success rate (ASR). Meanwhile,  we can observe that the detection performance via  {\explainer} on these  datasets are quite high,  especially for the nodes with {low degree} achieving around 0.4 under the NDCG metric. It means that the adversarial edges are highly likely to be ranked top among all edges which contribute to the model's prediction.
In other words, {\explainer} can generate a small subgraph (with some influential nodes) to reduce search space from millions of edges, and adversarial edges ranked top among all edges in the subgraph are likely to be inspected by domain experts. Thus, these observations indicate that {\explainer} has the potential to mark the adversarial edges in corrupted graph data for GNNs. \textbf{Note that similar observations on another representative explainer (PGExplainer~\cite{luo2020parameterized}) can be found in Figure~\ref{fig:inspector_citeseer} (Refer to Appendix Section)}.

\textbf{Problem Statement.} Given the node classification task, the attacker aims to attack a specific target node $v_i \in {V}_t$ by performing small perturbations on the graph $G = (\mathbf{A}, \mathbf{X})$ and obtains the corrupted graph ${\hat{G}} = (\mathbf{\hat{A}}, \mathbf{\hat{X}})$, such that the predicted label of the target node $v_i$  can be manipulated~\cite{zugner2018adversarial,wu2019adversarial}.
 
There are two main types of adversarial perturbations on the graph, including structure attacks to modify the adjacency matrix $\mathbf{A}$ and feature attacks to modify the feature matrix $\mathbf{X}$.
For ${simplicity}^{\ref{footnode:adding_edges}}$, we focus on the structure attacks, where attackers only add fake edges to connect the target nodes with others under certain perturbation budget $\Delta$. 
Note that a fixed perturbation budget $\Delta$ can be constrained as: $\|  \mathbf{E}' \| = \|\hat{ \mathbf{A}} - {\mathbf{A}}\|_{0} \leq\Delta$, 
 
where $\mathbf{E}'$ denotes the added adversarial edges by the attackers. In our work, in order to jointly attack a GNN model and its explanations, we design a new attacking method which is designed to achieve: (1) misleading the GNN model $f_\theta$ to give a wrong prediction $\hat{y}_i$ on node $v_i$; and (2) misleading the explanations of the GNN model such that the added fake edges do not appear in the output subgraph $\textbf{A}_S$ given by the {\explainer}~\cite{ying2019gnnexplainer}.
More formally, we state our attacking objective as:

\textbf{Problem: }\emph{Given $G=(\mathbf{ A}, \mathbf{ X})$, target (victim) nodes  $v_i \subseteq{ V}_t$ and specific target label  $\hat y_{i}$,  the attacker aims to select adversarial edges to composite a new graph $\hat{\mathbf{ A}}$ which fulfills the following two goals: (1)  The added adversarial edges can change the  GNN's prediction to a specific target label: $\hat y_i = \argmax_{c} f_\theta(\hat{\mathbf{A}}, \mathbf{X})_{v_i}^c$; and 
(2) The added adversarial edges will not be included in the subgraph generated by explainer: $  \hat{\mathbf{A}}-\mathbf{A} \notin { \mathbf{A}_S} $.
 }

\section{The Proposed Framework}
\label{sec:methodlogy}
In this section, we first introduce the basic component of graph attack via inserting adversarial edges and then propose how to bypass the detection of {\explainer}. Finally, we present the overall framework {\ourname} and the detailed algorithm.

\subsection{Graph Attack}
\label{sec:graph_attack}

We first formulate the problem to attack a GNN model as one optimization problem to search an optimal model structure $\hat{\mathbf{A}}$ which can let the model predict the victim node $v_i$ with a wrong label $\hat{y}_i$. In particular, given a well trained GNN model $f_{\theta}$ on the clean input graph $G$, we propose to achieve the attack on the target node $v_i$ with a specific target label $\hat y_i$ by searching $\hat{\mathbf{A}}$ which let the model have a minimum loss on $v_i$:
{\small
\begin{align}
  \min _{\hat{\mathbf{A}}}   \mathcal{L}_{\text{GNN}} (f_{\theta}(\hat{\mathbf{A}}, \mathbf{X})_{v_i}, \hat y_i)
  := - \sum^{C}_{c=1} \mathbb{I}[\hat y_i=c] \ln (f_{\theta}(\hat{\mathbf{A}}, \mathbf{X})_{v_i}^c ).
\end{align}
}\vspace*{-0.15\baselineskip}
Note that since we minimize the negative likelihood probability of the target label $\hat y_i$, the optimization of the above loss will promote the prediction probability of class $\hat y_i$ such that the prediction is maliciously manipulated from the original prediction $y_i$ to $\hat y_i$.  

To solve this optimization problem, we desire to apply the gradient-based attack methods, such as~\cite{goodfellow2014explaining} to figure out each input edge's influence on the model output. However, due  to  the  discrete  property  and  cascading  effects of graph data~\cite{zugner2018adversarial,jin2020adversarial}, it is not straightforward to attack a GNN model 
via the gradient-based attack method, like FGSM in continued space in computer vision tasks~\cite{goodfellow2014explaining}. 
To address the graph attack problem, we first relax the adjacency matrix $\mathbf{A} \in \{0,1\}^{n\times n}$ as continuous variable $\mathbb{R}^{n\times n}$. The calculated gradient information can help us approximately find the most ``adversarial'' edge in the current adjacency matrix, which is the element in the gradient that has the largest negative value. Once added this founded edge into the input graph, the model is highly likely to give a wrong prediction.

 \subsection{{\explainer} Attack}
 
The graph attack introduced in Section~\ref{sec:graph_attack} is usually satisfactory in terms of the successful attacking rate (ASR and ASR-T) as we will show in Section~\ref{sec:Experiments}.
However, just like existing attack methods on graph data, since the adversarial edges are highly correlated with the target prediction $\hat y_i$, they are most likely included in the subgraph generated by the {\explainer} and then become noticeable to the inspector or system designers.
Therefore, it is highly nontrivial to achieve attacking while bypassing the detection by {\explainer}. 

As introduced in Section~\ref{sec:pre}, {\explainer} aims to identify an important small subgraph $G_S$ that influences the GNN's prediction the most for making GNN's explanations. It works by minimizing $\mathcal{L}_{\text{Explainer}}$ (eq. ~\ref{equ: l_explainer_edges}) and selecting the top-$L$ edges in the adjacency mask matrix $\mathbf{M}_A$ with the largest values. 
Therefore, to bypass the detection by {\explainer}, we propose a novel \textit{{\explainer} attack} to suppress the possibility of adversarial edges being detected as follows:
\vspace*{-0.405\baselineskip}
{\small
\begin{align}
    \min_{\hat{\mathbf{A}} } \sum_{v_j\in \mathcal{N}(v_i)} \mathbf{M}_A^T[i,j] \cdot \mathbf{B}[i, j].
\end{align}
}\vspace*{-0.3\baselineskip}
\looseness=-1 
where ${\mathbf B} = \mathbf 1 \mathbf 1^T - \mathbf I - \mathbf{{A}}$.  $\mathbf I$ is an identity matrix, and $ \mathbf 1 \mathbf 1^T$ is all-ones matrix.  $ \mathbf 1 \mathbf 1^T - \mathbf I$ corresponds to the fully-connected graph.
When  $t$ is 0, $\mathbf{M}_A^{0}$ is randomly initialized; while $t$ is larger than 0, $\mathbf{M}_A^{t}$ is updated with step-size $\eta$ as follows:
{\small
\begin{align}
\mathbf{M}_A^{t} = 
\mathbf{M}_A^{t-1} - \eta \nabla_{\mathbf{M}_A^{t-1}} \mathcal{L}_{\text{Explainer}}(f_{\theta}, \hat{\mathbf{A}}, \mathbf{M}_A^{t-1}, \mathbf{X}, v_i, \hat y_i). 
\end{align}
} 
There are several key motivations:
 
 \begin{list}{\labelitemi}{\leftmargin=1em}
    \setlength{\topmargin}{0pt}
    \setlength{\itemsep}{0em}
    \setlength{\parskip}{0pt}
    \setlength{\parsep}{0pt}
    \item  
    The update of $\mathbf{M}_A^t$ mimics the gradient descent step in optimizing the loss function of {\explainer} in Eq.~\eqref{eq:explainer_loss} and $\mathbf{M}_A^T$ corresponds to the adjacency mask matrix after $T$ steps of update. 
    
    \item The loss term represents the total value of the adjacency mask corresponding to the edges between node $v_i$ and its direct neighbors $\mathcal{N}(v_i)$ since we focus on direct attack. Therefore, the adversarial edges we search among those neighbors tend to have a small value in the mask matrix $\mathbf{M}_A$; Since {\explainer} only selects edges with large values to construct the subgraph, there is a higher chance that adversarial edges could bypass the detection.
    
    \item 
    The penalty on existing edges in the clean graph is excluded by the matrix $\mathbf{B}$ where $\mathbf{B}[i,j]=0$ if edge $(v_i,v_j)$ exists in the clean graph $\mathbf{A}$. In this way, {\explainer} is still able to include normal edges in the subgraph. In other words, the {\explainer} works normally if not being attacked.
  \end{list}
Note that this loss function essentially accumulates and penalizes the gradient of $\mathcal{L}_{\text{Explainer}}$ with respect to $\mathbf{M}_A^t$ along the optimization path $\mathbf{M}_A^0 \rightarrow \mathbf{M}_A^1 \rightarrow \dots \rightarrow \mathbf{M}_A^T$. Each step of the gradient has a sophisticated dependency on the optimization variable $\hat{\mathbf{A}}$ and it requires the high-order gradient computation which is supported by deep learning frameworks such as PyTorch  and TensorFlow.

 \subsection{{\ourname}}
\label{sec:ourmodel}

After introducing the graph attack and {\explainer} attack, we finally obtain our proposed GEAttack framework as follows:
{\small
\begin{align}
    \min_{\hat{\mathbf{A}} }   \mathcal{L}_{\text{GEAttack}} :=  \mathcal{L}_{\text{GNN}} (f_{\theta}(\hat{\mathbf{A}}, \mathbf{X})_{v_i}, \hat y_i) + \lambda 
    \sum_{v_j \in \mathcal{N}(v_i)} \mathbf{M}_A^T[i,j] \cdot \mathbf{B}[i,j].
\end{align}
}
 
where $\mathbf{M}_A^{0}$ is randomly initialized when  $t$ is 0, and  for $t>0$, $\mathbf{M}_A^{t}$ can be updated as follows:
 {\small
\begin{align}
\mathbf{M}_A^{t} = 
\mathbf{M}_{A^{t-1}} - \eta \nabla_{\mathbf{M}_A^{t-1}} \mathcal{L}_{\text{Explainer}}(f_{\theta}, \hat{\mathbf{A}}, \mathbf{M}_A^{t-1}, \mathbf{X}, v_i, \hat y_i). 
\end{align}}
 The first loss term $\mathcal{L}_\text{GNN}$ guides the search of adversary edges such that the prediction of node $v_i$ is attacked; the second loss term guides the search process to bypass the detection of {\explainer}; and $\lambda$ is a hyperparameter which balances these two losses.
We propose \textbf{GEAttack} to solve this optimization problem as shown in Algorithm~\ref{alg: GEAttack}. It majorly runs two loops:
\begin{itemize}
\item In the inner loop, we mimic the optimization process of {\explainer} to obtain the adjacency mask $\mathbf{M}_A^T$ by $T$ steps of gradient descent. Note that we maintain the computation graph of these updates such that the dependency of $\mathbf{M}_A^T$ on $\hat{\mathbf{A}}$ is maintained, which facilitates the gradient computation in the outer loop;
\item In the outer loop, we compute the gradient of $\mathcal{L}_{\text{GEAttack}}$ with respect to $\hat{\mathbf{A}}$. Note that this step requires the backward propagation through all gradient descent updates in the inner loop and requires high-order gradient computation which is supported by the Automatic Differentiation Package in PyTorch and TensorFlow. In each iteration, we select one adversarial edge (set $\hat{\mathbf{A}} [i,j]=1$) according to the largest value in this gradient since this update will decrease the loss maximally, similar to the greedy coordinate descent algorithm.
\end{itemize}

{\small
\begin{algorithm}
 
\caption{ \textbf{{\ourname}}}
\begin{algorithmic}[1]
\State \textbf{Input}:  perturbation budget: $\Delta$; step-size and update iterations of {\explainer}: $\eta$, $T$; target node $v_i$; target label $\hat y_i$; graph $G =(\mathbf{A},\mathbf{X} )$, and a GNN model: $f_\theta$. 

\State \textbf{Output}: the adversarial adjacency matrix $\hat{\mathbf{A}}$.
 \State   ${\mathbf B} = \mathbf 1 \mathbf 1^T - \mathbf I - \mathbf{{A}}$,  $\hat{\mathbf A} = \mathbf A$, and randomly initialize $\mathbf{M}_{A}^0$;
 
\For{$o =  1, 2, \ldots,  \Delta $}  \gray{// outer loop over $\hat{\mathbf A}$;}    
        \For{$t =    1, 2, \ldots, T$ }  \gray{// inner loop over $\mathbf{M}_{A}^t$;}
             \State  compute 
            $\mathbf{P}^{t}=\nabla_{\mathbf{M}_A^{t-1}} \mathcal{L}_{\text{Explainer}}(f_{\theta}, \hat{\mathbf{A}}, \mathbf{M}_A^{t-1}, \mathbf{X}, v_i, \hat y_i)$;
            \State gradient descent: $\mathbf{M}_{A}^{t} = \mathbf{M}_{A}^{t-1} - \eta \thinspace  \mathbf{P}^{t}   $;
        \EndFor  
     \State compute the gradient w.r.t. $\hat{\mathbf A}$: $\mathbf{Q}^o = \nabla_{\hat{\mathbf A}} \mathcal{L}_{\text{GEAttack}}$;
     \State select the edge between node pair $(v_i , v_j)$ with the maximum element $\mathbf{Q}^o[i,j]$ as the adversarial edge, and update  $\hat{\mathbf A}[i,j] =1$ and $\mathbf{B}[i,j]=0$;
\EndFor
\State \textbf{Return} $\hat{\mathbf{A}}$.
\end{algorithmic}
\label{alg: GEAttack}
 \end{algorithm}

 }


\section{Experiment}
\label{sec:Experiments}
In this section, we conduct experiments to verify the effectiveness of our attacking model. 
We first introduce the experimental settings, then discuss the performance comparison results with various baselines, and finally study the effect of different model components on our model. Note that we also provide extended experimental settings (i.e., parameter settings in Appendix~\ref{sec:parameter_settings} and evaluation metrics in Appendix~\ref{sec:evaluation_metrics}) and results to show deep insights of GEAttack in Appendix Section.

 \subsection{Experimental Settings}
\label{sec:experiments_setting}
  \textbf{Datasets.} We conduct experiments on three widely used benchmark datasets for node classification, including CITESEER~\cite{kipf2016semi}, CORA~\cite{kipf2016semi}, and ACM~\cite{wang2020gcn,wang2019heterogeneous}. The processed datasets can be found in the github link\footnote{https://github.com/DSE-MSU/DeepRobust/tree/master/deeprobust/graph\label{github}}.
Following~\cite{zugner_adversarial_2019}, we only consider the largest connected component (LCC) of each graph. More details of datasets are provided in Appendix~\ref{sec:datasets}.

\textbf{Baselines.} Since the problem for jointly attacking GNN and {\explainer} in this paper is a novel task, there are no joint attack baselines. Thus, we mainly compare it with the state-of-the-art adversarial attack algorithms. We choose five baselines~\cite{li2020deeprobust}\textsuperscript{\ref{github}}, Random Attack (RNA), FGA~\cite{jin2020adversarial}, FGA-T, Nettack~\cite{zugner2018adversarial}, and IG-Attack~\cite{wu2019adversarial}. As most baselines are not directly applicable in target attack with a specific target label, we modify the attacking operations accordingly, such as modifying the  loss function with the specific target label, or constraining adversarial edges connecting with nodes who have the specific target label. Moreover, we also develop a straightforward baseline (\textbf{FGA-T$\&$E}) to jointly attack a GNN model and its explanations. More details of baselines are provided in Appendix~\ref{sec:add_baselines}.

\textbf{Attacker Settings.} In our experiments, we perform the target attack by selecting a set of target nodes under white-box setting and only consider the adding fake edges when doing adversarial perturbations.
Following the setting of IG-Attack~\cite{wu2019adversarial}, we select in total 40 victim target nodes which contain the 10 nodes with top scores, 10 nodes with the lowest scores, and the remaining nodes are randomly selected.
Note that we conducted direct attacks on the edges directly connected to the target node with a specific target label. 
To obtain a specific target label for each node, we first perform attack to fool the target nodes via the basic FGA attack method. The changed label for each target node then is set to be the specific target label if success. 
Note that we use these successfully attacked nodes to evaluate the final attacking performance. 
In addition, we conduct the evasion attack, where attacking happens after the GNN model is trained or in the test phase. The model is fixed, and the attacker cannot change the model parameter or structure.  The perturbation budget $\Delta$ of each target node is set to its degree.  Note that we report the average performance of 5 runs with standard deviations.

\begin{table*}[t]
 \centering
\caption{Results with standard deviations ($\pm$std) on three datasets using different attacking algorithms.
}
 \label{tab:comparsion_all}

\scalebox{0.800}
{
\begin{tabular}{c|c|c|c|c|c|c|c|c}
\hline
                   & \textbf{Metrics (\%)}      & \textbf{FGA}\footnote{} & \textbf{RNA} & \textbf{FGA-T} & \textbf{Nettack} & \textbf{IG-Attack} & \textbf{FGA-T$\&$E} & \textbf{{\ourname}} \\ \hline
\multirow{6}{*}{\rotatebox{90}{\textbf{CITERSEER}}} & \textbf{ASR}        & 86.79$\pm$0.08        & 55.52$\pm$0.08            & 99.56$\pm$0.01          & 99.11$\pm$0.01       & 91.54$\pm$0.05         & 98.74$\pm$0.02              & \textbf{100$\pm$0.00   }     \\ \cline{2-9} 
                                    & \textbf{ASR-T}      & -        & 54.27$\pm$0.10             & 99.56$\pm$0.01          & 99.11$\pm$0.01       & 91.54$\pm$0.05         & 98.74$\pm$0.02              & \textbf{100$\pm$0.00 }       \\ \cline{2-9}  
                                    & \textbf{Precision} & 13.45$\pm$0.01        & 9.96$\pm$0.01             & 13.44$\pm$0.02          & 10.21$\pm$0.01       & 10.21$\pm$0.01         & 13.31$\pm$0.01              & \textbf{9.87$\pm$0.02}    \\ \cline{2-9}  
                                    & \textbf{Recall}  & 74.55$\pm$0.05        & \textbf{63.80$\pm$0.05}             & 74.55$\pm$0.05          & 66.48$\pm$0.06       & 65.73$\pm$0.04         & 74.28$\pm$0.05              & 64.05$\pm$0.07   \\ \cline{2-9}  
                                    & \textbf{F1}      & 21.65$\pm$0.02        & \textbf{16.44$\pm$0.02 }           & 21.64$\pm$0.02          & 17.08$\pm$0.02       & 16.96$\pm$0.02         & 21.47$\pm$0.02              & 16.49$\pm$0.03   \\ \cline{2-9}   
                                    & \textbf{NDCG}    & 47.18$\pm$0.04        & 39.21$\pm$0.04            & 46.60$\pm$0.04           & 38.45$\pm$0.05       & 40.26$\pm$0.04         & 47.02$\pm$0.05              & \textbf{36.11$\pm$0.05}   \\ \hline \hline
\multirow{6}{*}{\rotatebox{90}{\textbf{CORA}}}     & \textbf{ASR}        & 90.54$\pm$0.05        & 62.97$\pm$0.10             & \textbf{100$\pm$0.00}               & \textbf{100$\pm$0.00}            & 90.17$\pm$0.07         & 99.79$\pm$0.01              & \textbf{100$\pm$0.00}        \\ \cline{2-9} 
                                    & \textbf{ASR-T}      & -        & 62.58$\pm$0.10             & \textbf{100$\pm$0.00}               & \textbf{100$\pm$0.00}            & 90.17$\pm$0.07         & 99.79$\pm$0.01              & \textbf{100$\pm$0.00}        \\ \cline{2-9}  
                                    & \textbf{Precision} & 16.02$\pm$0.01        & \textbf{10.47$\pm$0.01}            & 16.08$\pm$0.01          & 12.78$\pm$0.01       & 13.47$\pm$0.03         & 15.95$\pm$0.01              & 12.21$\pm$0.01   \\ \cline{2-9}  
                                    & \textbf{Recall}  & 72.65$\pm$0.05        & \textbf{55.40$\pm$0.07}             & 72.75$\pm$0.05          & 63.83$\pm$0.06       & 67.66$\pm$0.04         & 72.45$\pm$0.05              & 65.03$\pm$0.06   \\ \cline{2-9}  
                                    & \textbf{F1}      & 25.30$\pm$0.02         & \textbf{17.00$\pm$0.02 }              & 25.38$\pm$0.02          & 20.64$\pm$0.02       & 21.79$\pm$0.04         & 25.21$\pm$0.02              & 20.06$\pm$0.02   \\ \cline{2-9}  
                                    & \textbf{NDCG}    & 43.15$\pm$0.04        & \textbf{34.16$\pm$0.05 }           & 43.41$\pm$0.04          & 36.47$\pm$0.04       & 38.05$\pm$0.05         & 43.46$\pm$0.04              & 35.60$\pm$0.03    \\ \hline \hline
\multirow{6}{*}{\rotatebox{90}{\textbf{ACM}}}      & \textbf{ASR}        & 67.50$\pm$0.07         & 63.66$\pm$0.13            & \textbf{100$\pm$0.00}               & 98.00$\pm$0.03          & 98.82$\pm$0.02         & \textbf{100$\pm$0.00}                   & \textbf{100$\pm$0.00}        \\ \cline{2-9} 
                                    & \textbf{ASR-T}      & -         & 63.66$\pm$0.13            & \textbf{100$\pm$0.00}               & 98.00$\pm$0.03          & 98.82$\pm$0.02         & \textbf{100$\pm$0.00}                   & \textbf{100$\pm$0.00}        \\ \cline{2-9}  
                                    & \textbf{Precision} & 11.57$\pm$0.05        & \textbf{9.26$\pm$0.01}             & 11.88$\pm$0.05          & 12.98$\pm$0.03       & 11.69$\pm$0.05         & 11.31$\pm$0.05              & 9.61$\pm$0.02    \\ \cline{2-9}  
                                    & \textbf{Recall}  & 38.21$\pm$0.12        & \textbf{34.05$\pm$0.05}            & 38.34$\pm$0.12          & 43.67$\pm$0.09       & 44.49$\pm$0.14         & 37.90$\pm$0.12               & 38.08$\pm$0.08   \\ \cline{2-9}  
                                    & \textbf{F1}      & 14.16$\pm$0.05        & \textbf{12.75$\pm$0.02}            & 14.35$\pm$0.05          & 17.61$\pm$0.04       & 16.61$\pm$0.07         & 13.91$\pm$0.05              & 14.03$\pm$0.03   \\ \cline{2-9}  
                                    & \textbf{NDCG}    & 38.58$\pm$0.14        & 36.68$\pm$0.10             & 38.17$\pm$0.13          & 46.90$\pm$0.09        & 41.23$\pm$0.13         & 38.07$\pm$0.13              & \textbf{24.43$\pm$0.06}   \\ \hline
\end{tabular}
}
\begin{tablenotes}
   \item[] ~\textsuperscript{3} FGA cannot evaluate ASR-T metric where the specific target label are not available.
\end{tablenotes}
 
\end{table*}

 \subsection{Attack Performance Comparison}
\label{sec:comparison}
 We first evaluate how the attack methods perform and whether the adversarial edges can be detected by {\explainer}. The results are demonstrated in Table~\ref{tab:comparsion_all}. According to the results, we have the following observations.

 \textbf{Attacking GNN Model.} Our proposed attacker {\ourname} works consistently comparable to or outperform other strong GNN attacking methods. In all three datasets (CITESEER, CORA, and ACM), our proposed attacker {\ourname} achieves around 100\% attacking success rate when doing adversarial attacks with and without target labels (ASR-T \& ASR). It suggests that {\ourname} can achieve similar attacking power compared to other strongest GNN attackers such as FGA-T and Nettack, while also outperforming other attackers such as IG-Attack and random attack (RNA).

 \textbf{Attacking \explainer.} Our proposed attacker {\ourname} consistently outperforms other methods when attacking the \explainer, except for the RNA method.  In other words, our proposed {\ourname} is much harder to be detected by \explainer~ than all other attacking methods, only except for the RNA attacker. 
\emph{Note that the RNA method is the strongest baseline with regard to evade the detection of \explainer, while having the worst performance on attacking the GNN model with the ASR-T \& ASR metrics.} That is due to the fact that RNA attacker randomly adds edges to the target node, so the added edge is expected to have low influence on the model's prediction. 
From our experimental results, we could see, when excluding RNA attacker, our proposed {\ourname} is the most strongest attacker for \explainer. 
Compared to the most successful GNN attackers (Nettack and FGA-T), {\ourname} can let the \explainer~ have much lower Precision, Recall, F1, and NDCG score, which suggests that the \explainer~ has much lower power to detect adversarial perturbations from {\ourname}.
For another baseline method, FGA-T\&E which also tries to evade the \explainer~ (by considering only attack the edges that are not selected by \explainer), the \explainer~detector still has a high chance to figure out the adversarial perturbations.
In conclusion, our proposed {\ourname} can have good performance for attacking GNN models, which is comparable to other strongest attackers.  At the same time, it is much harder to be detected by \explainer. The experimental results can verify that our proposed method can jointly attack both a GNN model and its explanations ({\explainer}).

 \subsection{Jointly attacking GNNs and PGExplainer}
\label{sec:jointly_pge}
 
In this section, in order to evaluate the effectiveness of our proposed attacking method on both GNNs and its explanations, we apply our proposed method to another representative explainer for the GNNs model (PGExplainer~\cite{luo2020parameterized}), which adopts a deep model  to parameterize the generation process of explanations in the inductive setting. As shown in Figure~\ref{fig:inspector_citeseer} (Appendix Section),  we first conducted empirical studies to validate that PGExplainer has the potential to mark the adversarial edges in corrupted graph data for GNNs, which has similar observations on {\explainer} in Section~\ref{sec:pre}.

To perform jointly attacking, we adopt a similar manner to the search of adversarial edges via the gradient computation of PGExplainer.  
Table~\ref{tab:my-table_citeseer_PGE} shows the overall attack performance comparison on CITESEER dataset.  We do not show the results on CORA and ACM datasets since similar observations can be made. In general, we find that our proposed  attacker  GEAttack  achieves  the highest attacking success rate (ASR/ASR-T) compared with baselines. Meanwhile, as for attacking PGExplainer, our proposed attacker GEAttack also consistently outperforms other methods under Precision/Recall/F1/NDCG metrics when attacking the PGExplainer, except for the RNA method.  
Note that as RNA attacker randomly adds edges to the target node for jointly attacking, these adversarial edges might have a low influence to the model's prediction and could easily lead to evade the detection from Explainer, while making it difficult to attack the GNN model  under the ASR-T \& ASR metrics. These observations demonstrate that  both GNNs model and its explanations are vulnerable to adversarial attacks, and our proposed method can jointly attack both a GNN model and its explanations.

\begin{table*}[t]
\centering
 
\caption{{Results with standard deviations ($\pm$std) on CITESEER dataset using different attacking algorithms.}}

\label{tab:my-table_citeseer_PGE}
\scalebox{0.800}
{
\begin{tabular}{c|c|c|c|c|c|c|c}
\hline
\textbf{Metrics (\%)} & \textbf{FGA} & \textbf{RNA} & \textbf{FGA-T} & \textbf{Nettack} & \textbf{IG-Attack} & \textbf{FGA-T\&E} & \textbf{GEAttack} \\ \hline
\textbf{ASR}          & 88.89$\pm$0.06   & 55.19$\pm$0.04   & 99.24$\pm$0.01     & 97.20$\pm$0.18        & 98.93$\pm$0.01         & 98.76$\pm$0.01        & \textbf{99.34$\pm$0.03}        \\ \hline
\textbf{ASR-T}        & -            & 51.74$\pm$0.06   & 99.24$\pm$0.01     & 96.91$\pm$0.11       & 98.42$\pm$0.02         & 98.81$\pm$0.01        & \textbf{99.34$\pm$0.03}        \\ \hline
\textbf{Precision}   & 6.77$\pm$0.03    & 4.10$\pm$0.02     & 6.47$\pm$0.02      & 6.45$\pm$0.03        & 6.52$\pm$0.02          & 5.66$\pm$0.02         & \textbf{4.65$\pm$0.01}        \\ \hline
\textbf{Recall}    & 40.39$\pm$0.14   & 27.37$\pm$0.12   & 39.71$\pm$0.16     & 40.50$\pm$0.16        & 43.73$\pm$0.10          & 35.14$\pm$0.16        & \textbf{28.60$\pm$0.11}         \\ \hline
\textbf{F1}        & 11.07$\pm$0.04   & 6.79$\pm$0.03    & 10.61$\pm$0.04     & 10.65$\pm$0.05       & 10.72$\pm$0.03         & 9.19$\pm$0.04         & \textbf{7.47$\pm$0.02}         \\ \hline
\textbf{NDCG}      & 22.65$\pm$0.09   & 14.85$\pm$0.07   & 22.87$\pm$0.11     & 23.07$\pm$0.09       & 26.76$\pm$0.06         & 19.38$\pm$0.11        & \textbf{16.45$\pm$0.07}        \\ \hline
\end{tabular}
}
 
\end{table*}

\begin{wrapfigure}{R}{0.75\textwidth}
 \centering
{\subfigure[CORA - ASR-T]
{\includegraphics[width=0.3230\linewidth]{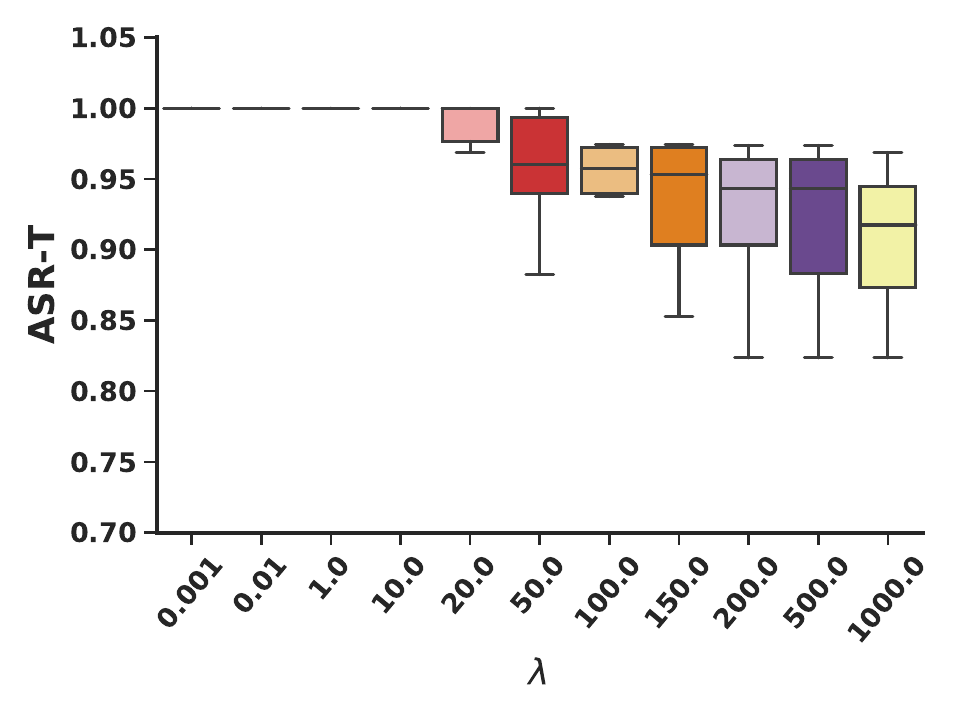}}}
{\subfigure[CORA - F1]
{\includegraphics[width=0.3230\linewidth]{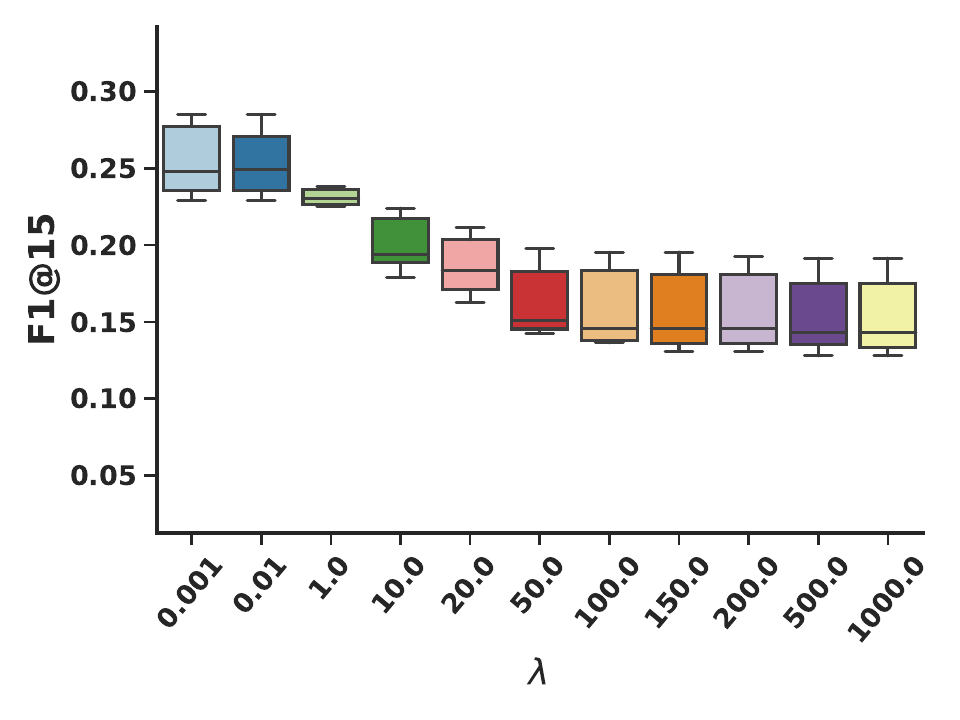}}}
{\subfigure[CORA - NDCG]
{\includegraphics[width=0.3230\linewidth]{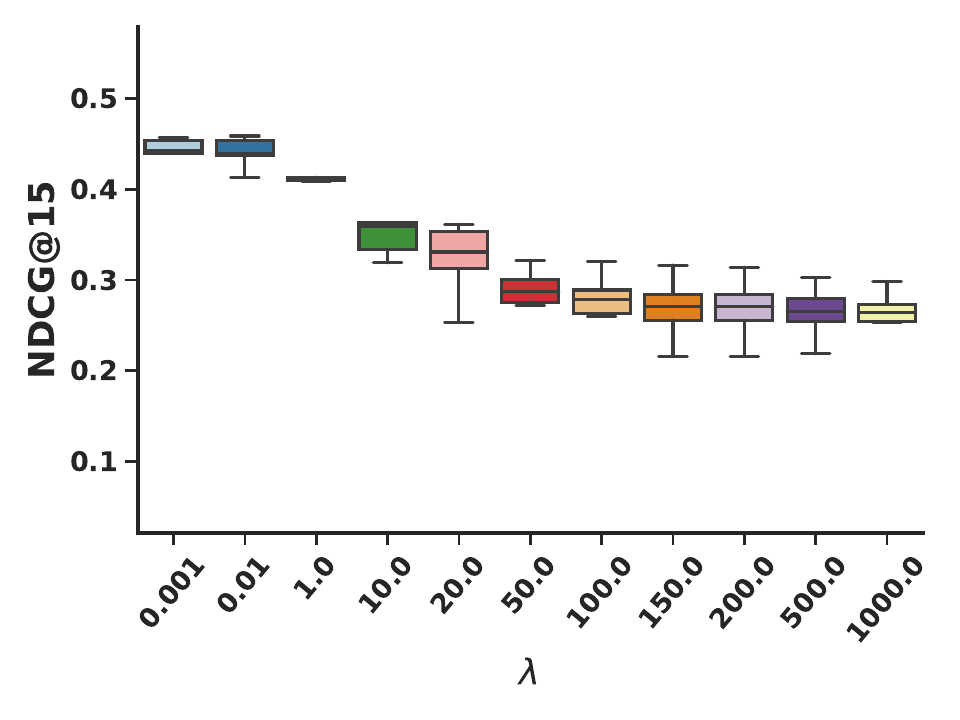}}}
\caption{Effect of  $\lambda$ under Attack Success Rate with Target label  (ASR-T) and detection rate (F1/NDCG) on  CORA dataset. 
}\label{fig:lambda_cora}  
 \end{wrapfigure}
  
 \subsection{Balancing the Graph Attack and {\explainer} Attack -  $\lambda$ }
 
In the previous subsection, we have demonstrated the effectiveness of the proposed method. In this subsection, we study the effect of model components between Graph Attack and {\explainer} Attack, which is controlled by $\lambda$. 
When $\lambda$ is close to 0, {\ourname} is degraded to Graph Attack model, while it focuses on {\explainer} Attack for larger values of $\lambda$.

The ASR-T performance change of {\ourname} on  CORA  dataset is illustrated in Figure~\ref{fig:lambda_cora}. As we can see from figures, the ASR-T of  {\ourname}  can maintain 100\% successfully attacked nodes  when $\lambda$ is set to 20.   
However, larger values of $\lambda$ can greatly hurt the  ASR-T performance. For instance, the ASR-T performance of {\ourname} can reduce to 95\% when $\lambda$ is set to 50. 
Moreover, from the figures, we first observe that when the value of $\lambda$ becomes large, the detection rate on  CORA dataset  consistently has the same trend under F1/NDCG metrics. In addition, the detection ratio maintains stable when the value of $\lambda$ is larger than 50. 
This observation suggests that a larger value of $\lambda$ is more likely to  encourage {\ourname} for selecting the adversarial edges as more unnoticeable as possible.

To summarize, larger values of  $\lambda$ can hurt Graph  Attack, while benefiting to $\explainer$ Attack, and vise versa.
These observations demonstrate that there may indeed exist the trade-off relation between attacking GNN model and the $\explainer$. However, selecting a proper $\lambda$ can facilitate us to achieve good attacking performance for the two adversarial goals simultaneously.
 Note that more results on CITESEER dataset regarding the effect of $\lambda$ are shown in Figure~\ref{fig:lambda_top15_citeseer} (Appendix Section).

 \section{Conclusion}
\label{sec:conclusion}

 In this paper, we first dispatched empirical studies to demonstrate that {\explainer}  can act as an inspection tool and have the potential to  detect the adversarial perturbations for graph data.  After that,  we introduced a new problem: \textit{Whether a graph neural network and its explanations can be jointly attacked by modifying graph data with malicious desires?} 
To address this problem, we  presented a novel attacking method (\textbf{{\ourname}})  to jointly attack a graph neural network and its explanations.
Our thorough experiments on several real-world datasets suggested the superiority of the proposed {\ourname} over a set of competitive baselines. Then, we furthermore performed the model analysis to better understand the behavior of {\ourname}. 

Currently, we only consider detecting adversarial edges via {\explainer}, while there exist other adversarial perturbations, like modifying features and injecting fake nodes. In the future, we would like to extend the proposed model for performing attacks via other types of adversarial perturbations. Moreover, we would like to extend the proposed framework on more complicated graph data such as heterogeneous information and dynamic graphs.

\bibliographystyle{unsrt}

\bibliography{references}

\newpage

 \appendix
 \begin{center}
    {\LARGE Supplementary Material: Jointly Attacking Graph Neural Network and its Explanations}
 \end{center}

\begin{figure*}[htbp]
 
\centering
{\subfigure[CITESEER - ASR]
{\includegraphics[width=0.316\linewidth]{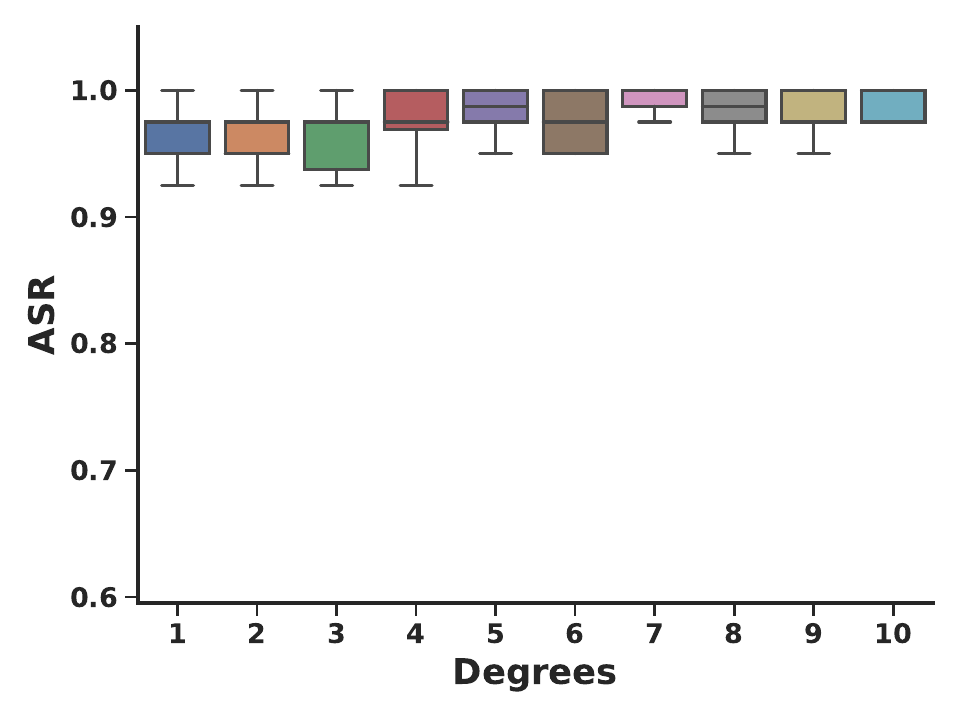}}}
{\subfigure[CITESEER - F1]
{\includegraphics[width=0.315\linewidth]{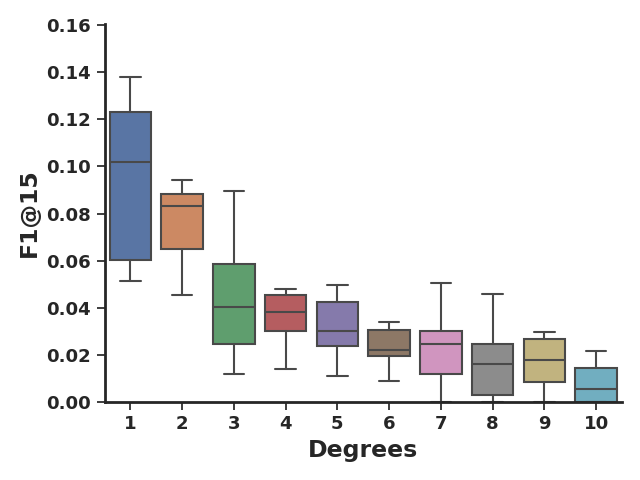}}}
{\subfigure[CITESEER - NDCG]
{\includegraphics[width=0.315\linewidth]{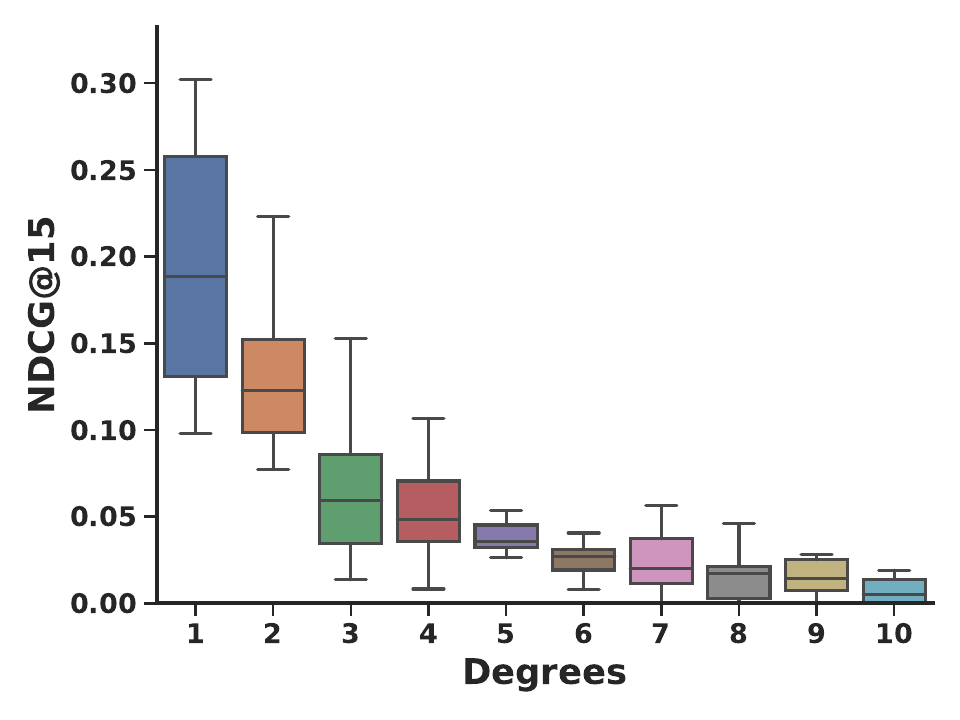}}}
{\subfigure[CORA - ASR]
{\includegraphics[width=0.316\linewidth]{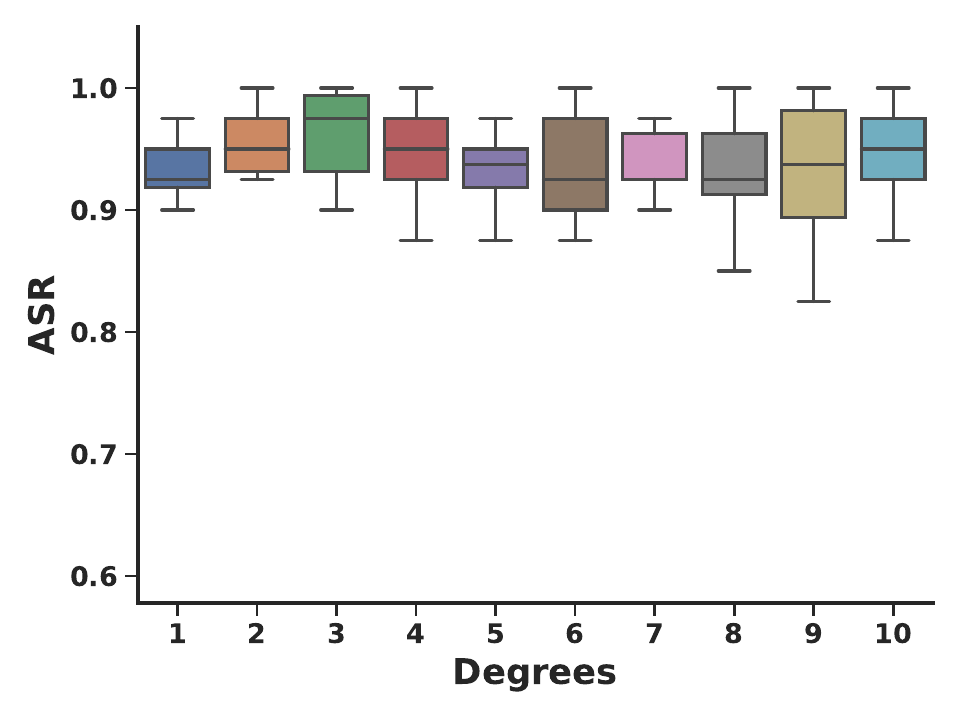}}}
{\subfigure[CORA - F1]
{\includegraphics[width=0.316\linewidth]{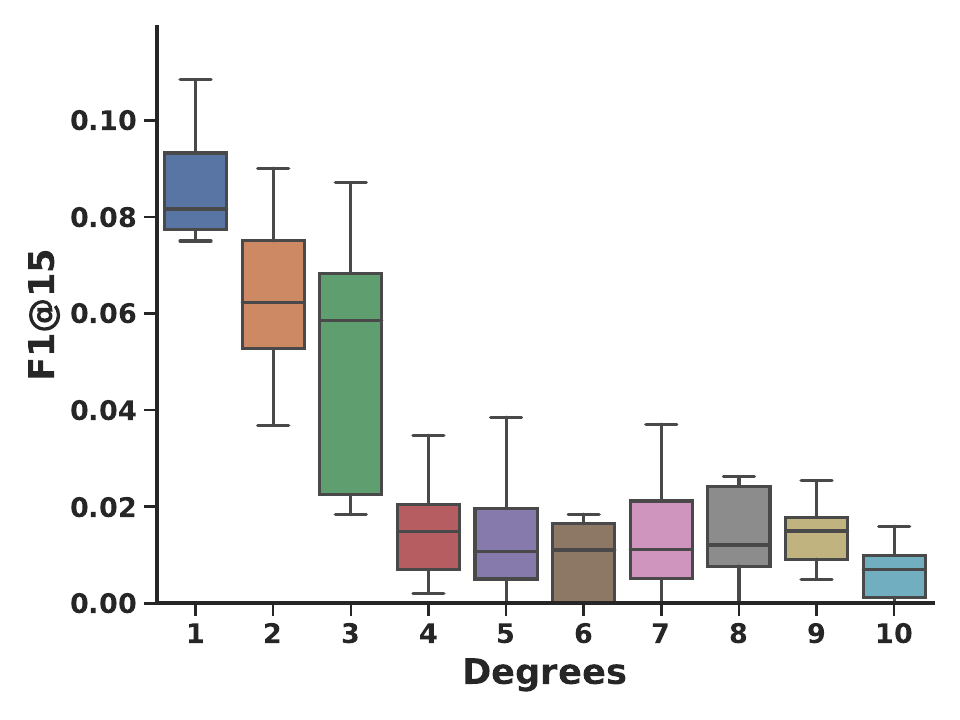}}}
{\subfigure[CORA - NDCG]
{\includegraphics[width=0.316\linewidth]{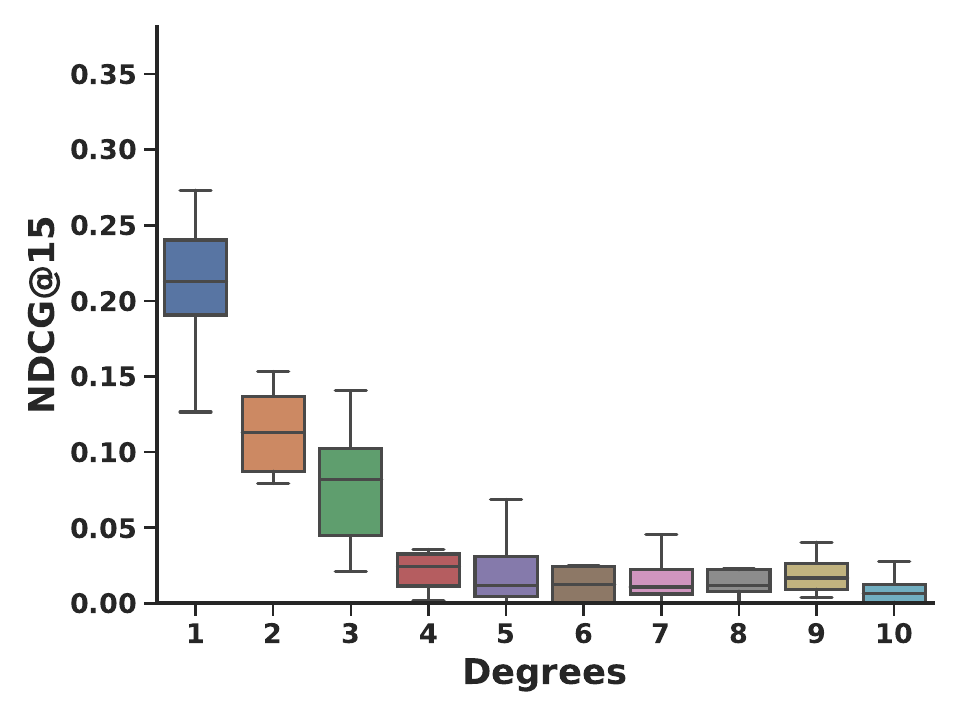}}}
\vskip -0.1in
\caption{Results of detecting the adversarial edges via PGExplainer Inspector under Nettack on CITESEER and CORA datasets.}\label{fig:inspector_citeseer}
\vskip -0.1in
\end{figure*}

\begin{figure*}[bp]
\centering
{\subfigure[CITESEER - Precision]
{\includegraphics[width=0.45\linewidth]{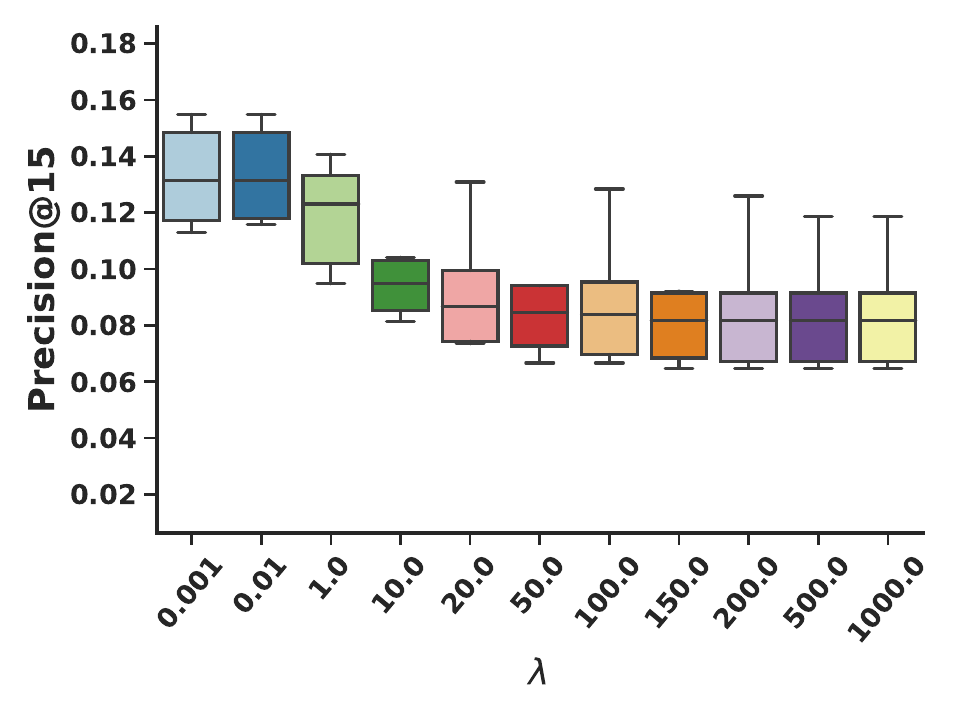}}}
{\subfigure[CITESEER - Recall]
{\includegraphics[width=0.45\linewidth]{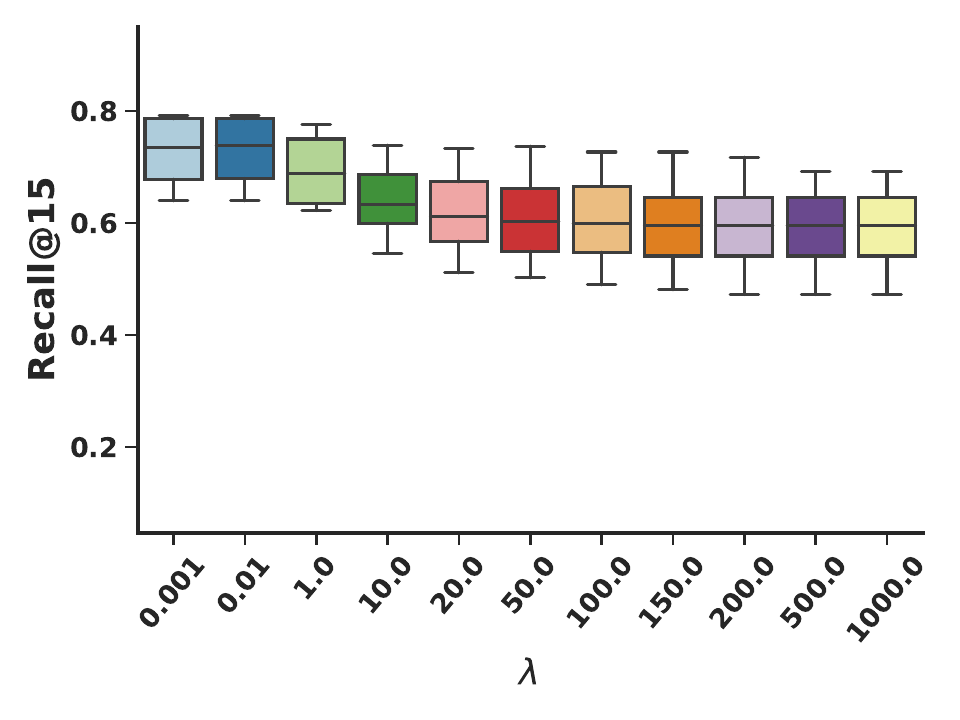}}}
{\subfigure[CITESEER - F1]
{\includegraphics[width=0.45\linewidth]{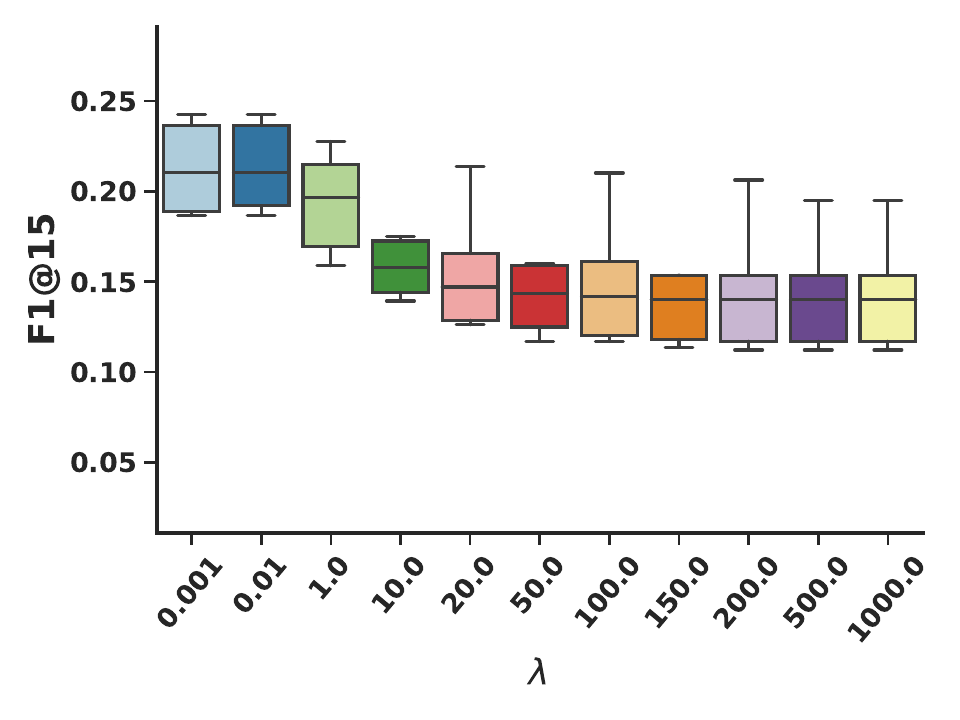}}}
{\subfigure[CITESEER - NDCG]
{\includegraphics[width=0.45\linewidth]{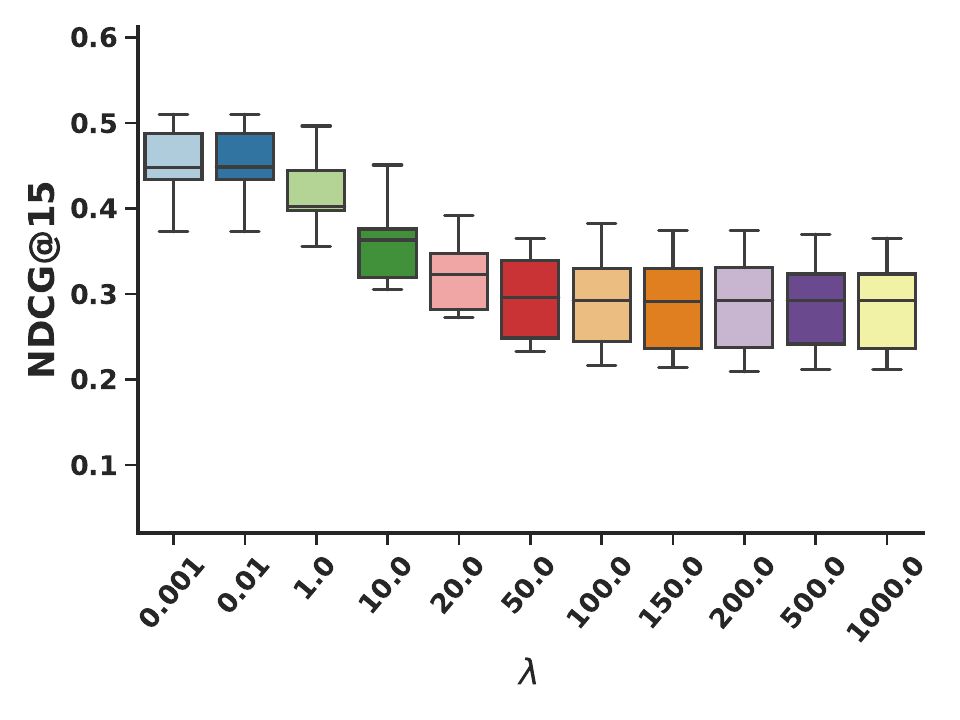}}}
\caption{Effect of   $\lambda$ under detection rate (Precision/Recall/F1/NDCG)  on CITESEER  dataset.}\label{fig:lambda_top15_citeseer}
\end{figure*}

In this section, we provide the necessary information for reproducing our insights and experimental results. 
These include  the detailed description of parameter settings in Section~\ref{sec:parameter_settings}, evaluation metrics in Section~\ref{sec:evaluation_metrics}, datasets in Section~\ref{sec:datasets}, the quantitative results on PGExplainer that can further support our insights in   Figure~\ref{fig:inspector_citeseer}, and hyper-parameters (i.e.,  $\lambda$, $T$, $L$) studies as shown in Figure~\ref{fig:subgraph_top15_cora},~\ref{fig:t_top10_15_acm},~\ref{fig:lambda_top15_citeseer}.

\section{Experimental Settings}
\label{sec:add_experimental_settings}

\subsection{\textbf{Parameter Settings}}  
\label{sec:parameter_settings}

For training the GNN model in each graph, we randomly choose 10\% of nodes for training, 10\% of nodes for validation and the remaining 80\% of nodes for test~\cite{jin2020graph}. The hyper-parameters of all the models are tuned based on the loss and accuracy on the validation set. 
Without any specific mention, we adopt the default parameter setting in the author's implementation.  
The implementation of our proposed method is based on the DeepRobust repository~\cite{li2020deeprobust}\textsuperscript{\ref{github2}}, a PyTorch library for adversarial attacks. 
The search spaces for hyper-parameters are as follows:
\begin{itemize}
\item $\lambda =\{0.001, 0.01, 1, 10, 20, 50, 100, 200,  500\}$
\item $d =\{4, 8, 16, 32, 64, 128\}$
\item $T =\{1, 2, 3, 4, 5, 6, 7, 8, 9, 10\}$
\item $L =\{5, 10, 20, 40, 60, 80, 100\}$
\item $\text{Learning rate}=\{0.01, 0.005, 0.001, 0.0005, 0.0001, 0.00005, 0.00001\}$
\end{itemize}

\subsection{Evaluation Metrics}
\label{sec:evaluation_metrics}

\textbf{Evaluation Metrics.} We evaluate the effectiveness of different attacking methods from two perspectives.  One type includes \textbf{Attack Success Rate (ASR)} ~\cite{ma2019attacking} and  \textbf{Attack Success Rate with Target label (ASR-T)}, which are the ratio of the successfully attacked nodes among all target nodes to any wrong label and specific (incorrect) target label. In our preliminary study in Section~\ref{sec:pre}, we have demonstrated that {\explainer} can act as an inspector for adversarial edges. Therefore, the other type of evaluation metrics are the popular accuracy metrics for detection rate~\cite{han2011data}: \textbf{Precision@K, Recall@K, F1@K}, and Normalized Discounted Cumulative Gain (\textbf{NDCG@K}). The first three metrics (Precision@K,  Recall@K, F1@K) focus on how many adversarial edges are included in the Top-K list of the subgraph generated via {\explainer}, while the last metric  (NDCG@K) accounts for the ranked position of adversarial edges in the Top-K list. We set $K$ as 15 only.  Note that adversarial edges with higher important weights in masked adjacency ($\mathbf{M}_{A}$) are more likely to present at top ranks and be easily detected by people (such as system inspectors or designers).  Hence, higher values of these metrics (Precision@K, Recall@K, F1@K, and NDCG@K) indicate that the adversarial edges are more likely to be detected and noticeable. Meanwhile, lower values of them indicate that adversarial edges are less likely to present in the subgraph ($G_S$) and more unnoticeable to human, where the {\explainer} can be attacked. Without any specific mention, we adopt the default parameter setting of {\explainer} in the author's implementation\footnote{https://github.com/RexYing/gnn-model-explainer}, and the size of subgraph $L$ is set to 20. Note that we further analyse the impact of {\explainer} inspector on adversarial edges based on the various size of subgraph $L$ at Section~\ref{sec:subgraph_size}.

 \subsection{Datasets}
\label{sec:datasets}

\begin{table}[h]
 \centering
\caption{The statistics of the datasets by considering the Largest Connected Component (LCC).}
 \scalebox{1.00}{
\begin{tabular}{c|cccc}
\toprule
\textbf{Datasets}         &  \textbf{Nodes} &  \textbf{Edges}  & \textbf{Classes} & \textbf{Features} \\ \midrule
CITESEER & 2,110 & 3,668  & 6       & 3,703     \\ 
CORA     & 2,485 & 5,069  & 7       & 1,433     \\ 
ACM      & 3,025 & 13,128 & 3       & 1,870        \\  \bottomrule
\end{tabular}
} \label{tab:dataset}
\end{table}

We conduct experiments on three widely used benchmark datasets for node classification, including CITESEER~\cite{kipf2016semi}, CORA~\cite{kipf2016semi}, and ACM~\cite{wang2020gcn,wang2019heterogeneous}. These processed datasets can be found in the github link\footnote{https://github.com/DSE- MSU/DeepRobust/tree/master/deeprobust/graph\label{github2}}.
The statistics of these three datasets are presented in Table~\ref{tab:dataset}.

\begin{itemize}
\item  \textbf{CITESEER}~\cite{kipf2016semi}. CITESEER is a research paper citation network with nodes representing papers and edges representing their citation relationship. The node labels are based on the paper topics and the node attributes are bag-of-words descriptions about the papers. 
\item  \textbf{CORA}~\cite{kipf2016semi}. CORA is also a citation network where nodes are papers and edges are the citation relationship between the papers. The node attributes are also bag-of-words descriptions about the papers.  The papers are divided into seven classes.
\item \textbf{ACM}~\cite{wang2020gcn}. This network is extracted from 
ACM dataset where nodes represent papers with bag-of-words representations as node attributes. The existence of an edge between two nodes indicates they are from the same author. The nodes are divided into three classes.  

\end{itemize}

\subsection{Additional Details on Baselines}
\label{sec:add_baselines}

The details for baselines are as follows:

\begin{itemize}
\item \textbf{Random Attack (RNA)}: The attacker randomly adds adversarial edges to connect the target node with one from candidate nodes whose label is specific target label until reaching the perturbation budget.

\item \textbf{FGA}~\cite{jin2020adversarial}: This is a gradient-based attack method which aims to find adversarial edges by calculating the gradient of model's output on the adjacency matrix. Note that this method does not consider to fool the model to specific label.  
\item \textbf{FGA-T}: Similar to FGA attack, FGA-T is a targeted version of FGA attack which aims to attack the target node to specific target label.

\item \textbf{Nettack}~\cite{zugner2018adversarial}: This method introduces the first study of adversarial attacks on graph data by  preserving important graph characteristics.

\item \textbf{IG-Attack}~\cite{wu2019adversarial}: This baseline introduces  an integrated gradients method that could accurately reflect the effect of perturbing edges for adversarial attacks on graph data.

\item \textbf{FGA-T$\&$E}: Another baseline based on FGA-T method, but further incorporates the desire to evade the detection {\explainer} when generating adversarial edges. We first adopt {\explainer} to generated a small subgraph. Then, we   exclude the potential nodes from the subgraph when generating the adversarial edges between the target node and the potential nodes.
   
\end{itemize}

\section{Additional Experiments}
\label{sec:supplemenation}

\begin{figure}[h]
 
\centering
{\subfigure[CORA - Precision]
{\includegraphics[width=0.24\linewidth]{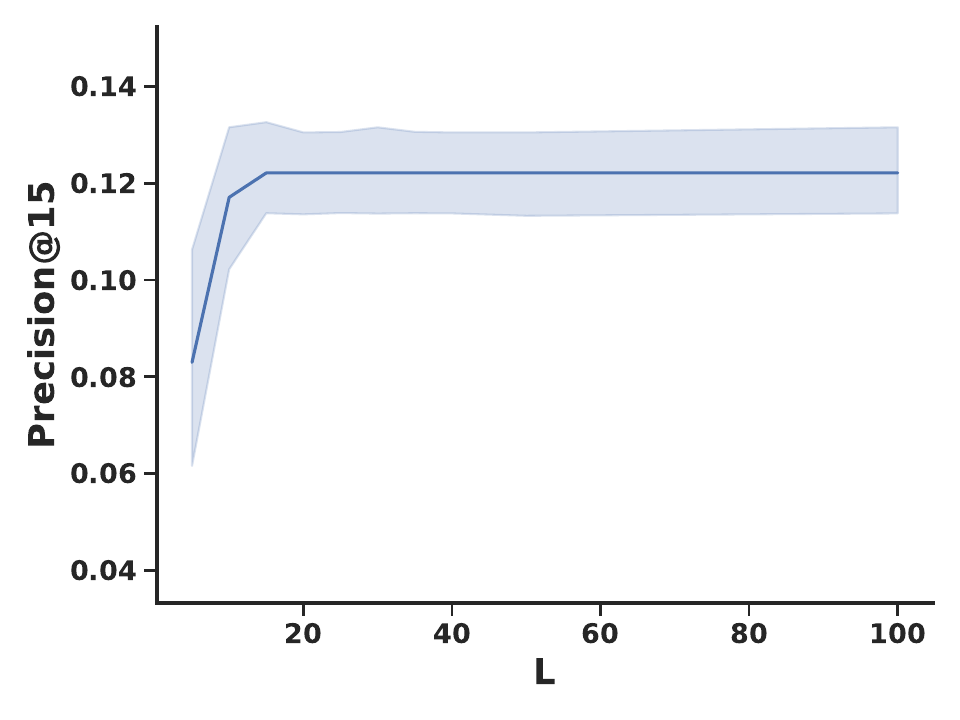}}}
{\subfigure[CORA - Recall]
{\includegraphics[width=0.24\linewidth]{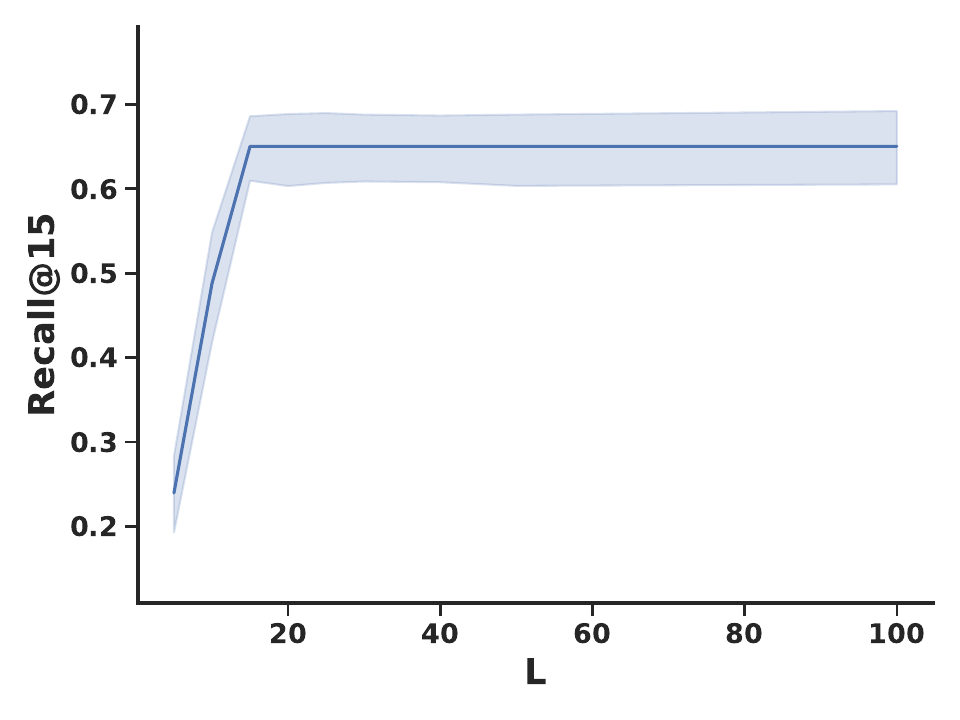}}}
{\subfigure[CORA - F1]
{\includegraphics[width=0.24\linewidth]{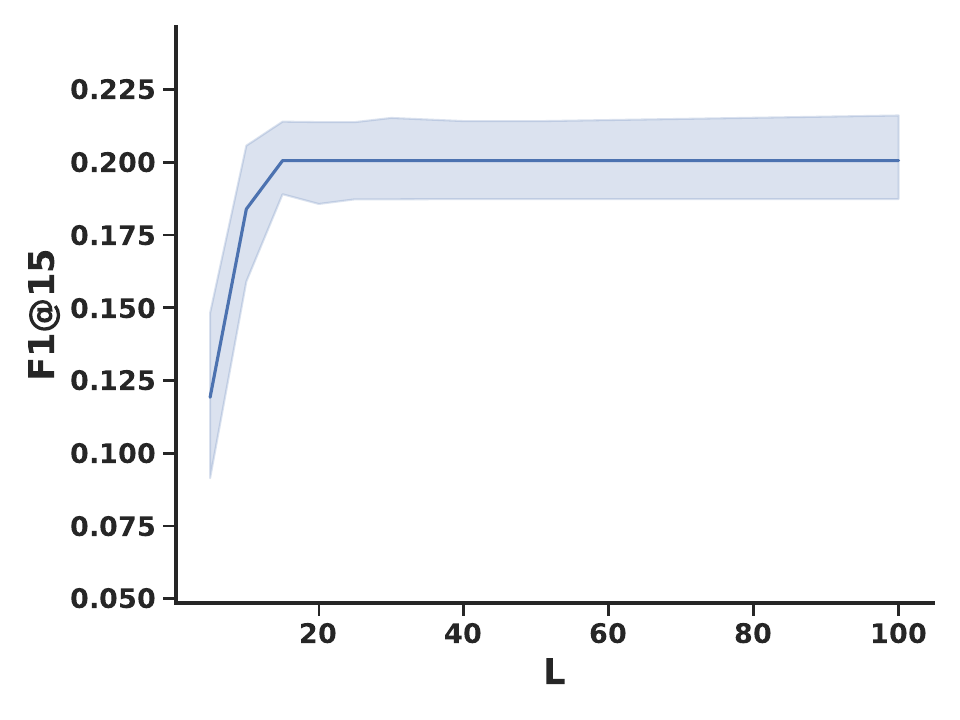}}}
{\subfigure[CORA - NDCG]
{\includegraphics[width=0.24\linewidth]{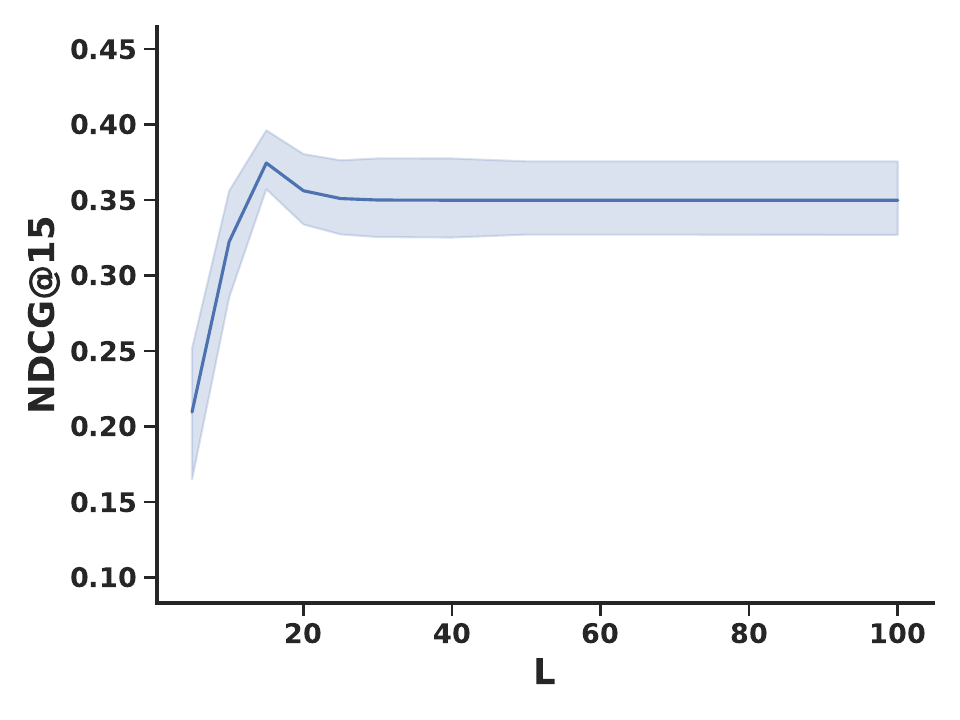}}}
\vskip -0.12in
\caption{Effect of  size of subgraph $L$ under  detection rate  (Precision/Recall/F1/NDCG) on   CORA  dataset.}\label{fig:subgraph_top15_cora}
 \end{figure}

\begin{figure}[t]
\centering
{\subfigure[CORA - F1]
{\includegraphics[width=0.24\linewidth]{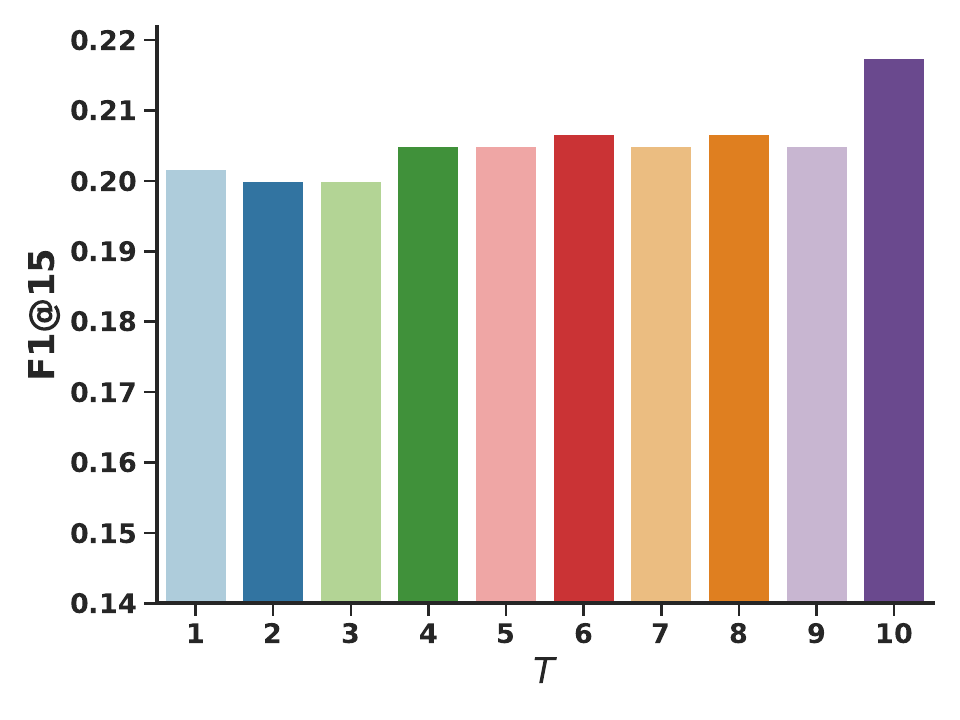}}}
{\subfigure[CORA - NDCG]
{\includegraphics[width=0.24\linewidth]{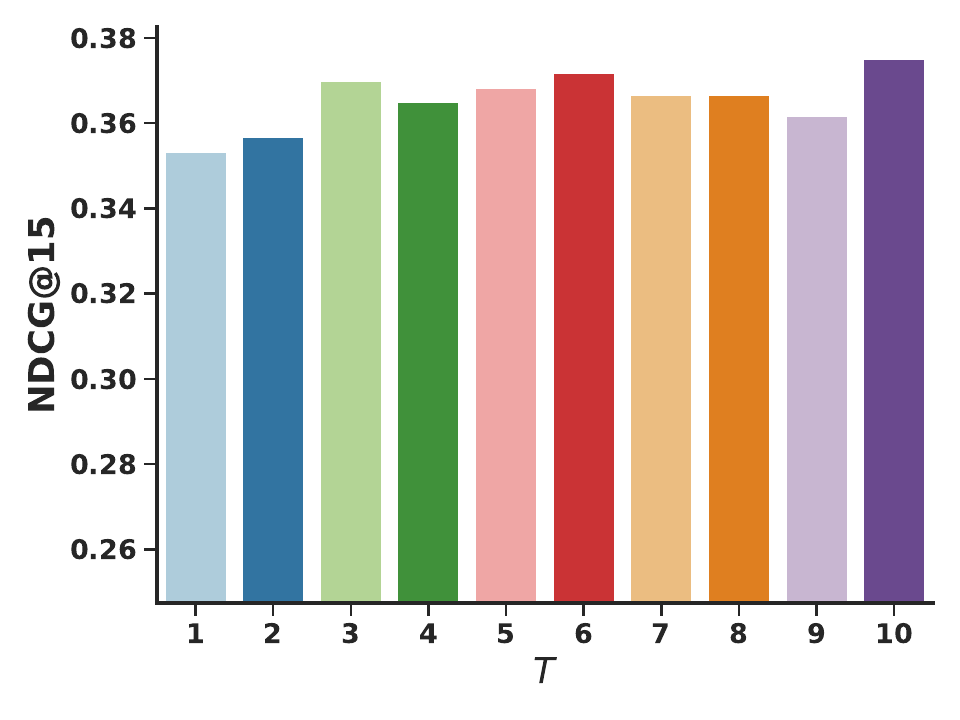}}}
{\subfigure[ACM - F1]
{\includegraphics[width=0.24\linewidth]{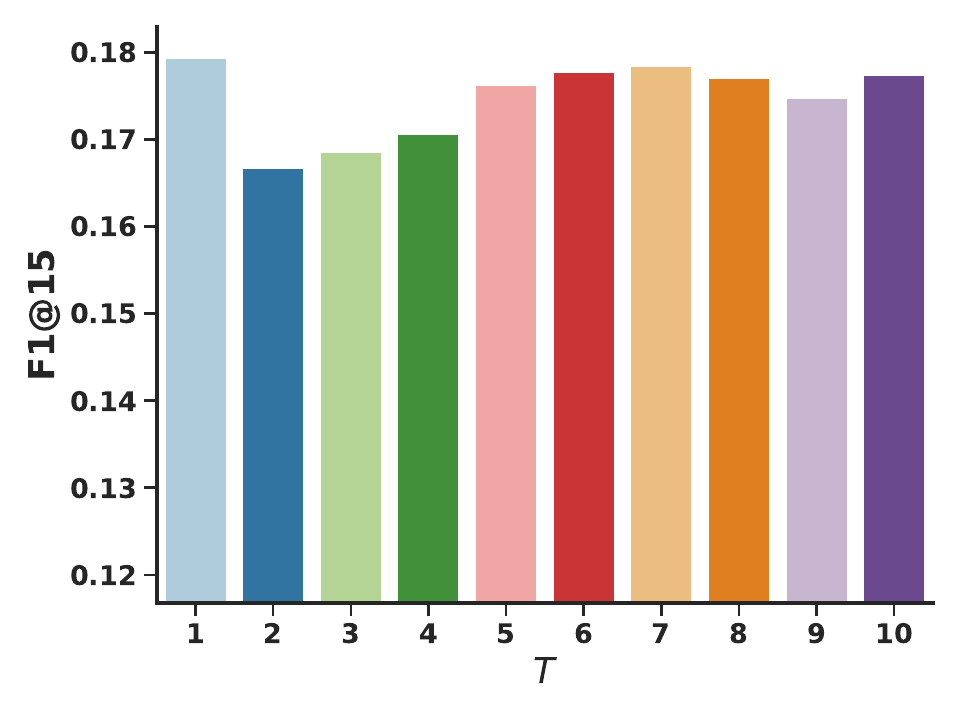}}}
{\subfigure[ACM - NDCG]
{\includegraphics[width=0.24\linewidth]{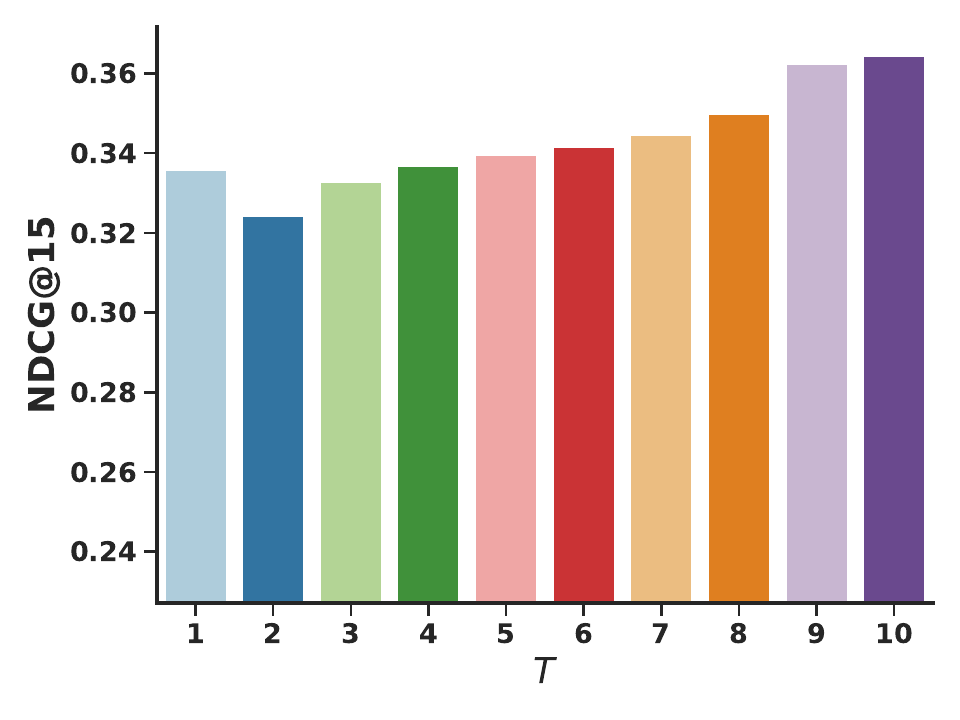}}}
 \caption{Effect of  $T$ under  detection rate  (F1/NDCG) on CORA and ACM datasets.}\label{fig:t_top10_15_acm}
 \end{figure}

\subsection{Parameter Analysis}
 In this section, we study the effect of model hyper-parameters for understanding the proposed method, including the size of subgraph $L$ and the number of update iterations $T$.  
\subsubsection{\textbf{Effect of  Subgraph Size $L$}}  
\label{sec:subgraph_size}
 
In this subsection, we further study the impact of  {\explainer} inspector for adversarial edges based on the size of subgraph $L$.  
Figure~\ref{fig:subgraph_top15_cora} shows the detection rate of {\ourname} with varied sizes of the subgraph $L$. 
As we can see, when the size of subgraph increases, the performance tends to increase first. And   {\ourname}  can not keep increasing when the size of subgraph is larger than around 20.  

\subsubsection{\textbf{Effect of the Number of Update Iterations $T$}} 

In this subsection, we explore the sensitivity of hyper-parameter $T$ for our proposed {\ourname}.  $T$  is the number of steps of updating  {\explainer}, which may influence the learning of $\mathbf{M}_{A}^t$.    The results are given in Figure ~\ref{fig:t_top10_15_acm} on CORA and ACM datasets. 
We do not show the results under attack success rate (ASR-T) as the performance almost achieves  100\% and do not change too much.
From the figure, we can observe that  our proposed {\ourname} method can achieve good performance under a small value of $T$ (i.e., less than 3), which indicates that  sub-optimal solutions of {\explainer} can provide sufficient gradient signal regarding $\mathbf{M}_{A}^t$  to guide the selection of adversarial edges for jointly attacking graph neural networks and {\explainer}.

\end{document}